\documentclass[twoside,11pt]{article}
\usepackage{jair, theapa, rawfonts}
\usepackage{graphicx}
\graphicspath{{./Images/}}
\DeclareGraphicsExtensions{.pdf,.jpg,.png,.eps}
\usepackage{amsmath}
\usepackage{float}
\restylefloat{table}
\usepackage[maxfloats=25]{morefloats}
\usepackage[caption=false]{subfig}
\usepackage{multicol, blindtext}
\usepackage{color,soul}
\usepackage[dvipsnames]{xcolor}
\usepackage{lipsum}
\usepackage{upgreek}
\usepackage{tablefootnote}
\usepackage{threeparttable}
\usepackage{array}
\usepackage{multirow}
\usepackage[center]{caption}
\usepackage[utf8]{inputenc}
\usepackage{amsmath,amssymb,amsfonts}%
\usepackage{amsthm}%
\usepackage{mathrsfs}%
\usepackage[title]{appendix}%
\usepackage{xcolor}%
\usepackage{textcomp}%
\usepackage{manyfoot}%
\usepackage{booktabs}%
\usepackage{algorithm}%
\usepackage{algorithmicx}%
\usepackage{algpseudocode}%
\usepackage{listings}%
\usepackage{algorithm}
\usepackage{algpseudocode}
\usepackage{algorithmicx}
\usepackage{tikz}
\usepackage{tkz-euclide}
\usepackage{upquote}
\usetikzlibrary{babel}
\usepackage{placeins}
\usepackage{pgfplots}
\usepgfplotslibrary{polar}
\usepgflibrary{shapes.geometric}
\usetikzlibrary{calc,angles,positioning,intersections,automata,arrows}
\algrenewcommand\textproc{}
\algdef{SE}[DOWHILE]{Do}{doWhile}{\algorithmicdo}[1]{\algorithmicwhile\ #1}
\UseRawInputEncoding

\ShortHeadings{Novelty Search Based Particle Swarm Optimization}
{Misra \& Kumar}
\firstpageno{1}

\begin{document}

\title{Particle Swarm Optimization based on Novelty Search}

\author{\name Rajesh Misra \email rajeshmisra.85@gmail.com \\
        \addr Department of Computer Science,\\ Seth Anandram Jaipuria College,10 Raja Naba Krishna Street,\\Kolkata,West Bengal,India\\
        \name Kumar Sankar Ray \email kumarsankar.ray@gla.ac.in\\
        \addr Department of Computer Science,\\ GLA University,\\Mathura,Uttar Pradesh,India\\   
       }


\maketitle

\begin{abstract}
In this paper we propose a Particle Swarm Optimization algorithm combined with Novelty Search. Novelty Search finds novel place to search in the search domain and then Particle Swarm Optimization rigorously searches that area for global optimum solution. This method is never blocked in local optima because it is controlled by Novelty Search which is objective free. For those functions where there are many more local optima and second global optimum is far from true optimum, the present method works successfully. The present algorithm never stops until it searches entire search area. A series of experimental trials prove the robustness and  effectiveness of the present algorithm on complex optimization test functions.
\end{abstract}

\section{Introduction}
\label{Introduction}

Particle Swarm Optimization has a long history of approximately more than two decades\\ \cite{Ken95}. A ton of variants surrounded in Swarm Optimization domain with various improvements and applications which often confuse researchers which algorithm to use and when. Maurice Clerc proposed a variant of PSO in 2011 which provides a standard for the class of particle swarm optimization\cite{Cle11} known as SPSO11. The SPSO11\cite{Cle11} helps researchers to compare their PSO variants in terms of performance. In 2013 when Liang et.al \cite{lia13} formalized a strong benchmark function on Real-Parameter Optimization problems. Swarm intelligence researchers dive into those benchmark suite to test the strength of their PSO variants.Clerc et .al also tested SPSO11 with that benchmark suite and provided a baseline for further improvements of PSO problems \cite{Zam13}.

Though SPSO11 is an earlier algorithm but it performs very well on most of the objective test functions. On the other hands it performs poor in non-separable and asymmetrical functions. The guideline for future PSO improvements is that new PSO algorithms must have the \textit{ability to solve non-separable and asymmetrical functions, with a large number of local minima and a second global minimum located far from the true optimum.}

We compare our algorithm with two very related works - one is \textit{Comprehensive Learning Particle Swarm Optimizer For Global Optimization of Multimodal Functions(CLPSO)}\\ \cite{Jq06} and other one is \textit{Enhanced comprehensive Learning Particle Swarm Optimization(ECLPSO)} \cite{Xi14}.We chose these two works because these two works are more relevant to our work. The above mentioned works mainly deal with multi-modal and composition objective test function, which we also perform. We choose those optimization functions where optimal values are more diverse and contain high number of false local optima. 

Lehman and Stanley proposed the concept of \textit{Novelty Search} in 2008, where they formulated an idea of objective-free search \cite{Leh08}. The search operation is based on behavioral novelty. They kept on showing the effectiveness of Novelty search on various other research works\cite{Leh10}\cite{Leh10b}. They showed how open-endness and objective free based search is quite effective and open a new way of thinking about unknown domain search\cite{Leh11}. Doncieux et. al provided theoretical explanation of Novelty search in \cite{Don19}.

The major problem of PSO and its other variants is that they \textit{stuck in local optima} which leads to premature convergence. Once a promising solution found by one particle of a swarm others gradually jump towards that location and create stagnation. It does not search the entire area of the search region finding nearby solution of local or global optimum. This is major concern of Particle Swarm Optimization algorithms and this is one of the reasons that PSO fails on those problems where diversity in searching is required. Using Novelty Search we can easily avoid this problem. Once a novel area of searching is found we apply PSO on that region to find the global optimum solution. Then again we search another novel area and apply PSO there. The search will not stop until there are no novel spaces are left. In this way the present algorithm never stuck in any local optima because the algorithm is never driven by best solution so far; rather it is driven by objective free Novelty search. 

In the present algorithm, Novelty Search controls PSO. First novelty search algorithm finds novel area for searching and then calls PSO to search that area for global optimum solution. We keep on continue this procedure until no further novel area is left. This is where the present algorithm differs from other PSO algorithms. Other PSO algorithms driven by objective; but the present algorithm is driven by novelty search which is objective free.

What advantages we can get by combining Novelty Search with Particle Swarm Optimization? Novelty search only check weather the search area is novel or not. It is not driven by objective. That is why it never gets stuck into some location rather it freely moves around the entire search space. The main cause of premature convergence is lack of divergence. After the population converges, it becomes very uniform (all solutions resemble the best one).But as Novelty search algorithm based on divergence, more diverge means higher probability of getting selected for PSO search. Novelty search helps to roam entire search space more quickly than any other evolutionary algorithm. We consider that advantage in PSO to make it more explorative algorithm.

The paper is organized as follows: Section 2 provides a brief introduction to novelty search. The basic concepts, general framework of our approach, Novelty computation of a search region  and novelty score calculation are discussed section 3. Section 4 provides the algorithmic summary and pseducode of the proposed Novelty Search based PSO algorithm.Section 5 provides detail experimental studies on benchmark datasets CEC-13 and CEC-20 and also with advanced ambiguous benchmark functions . Section 6 provides convergence divergence analysis of Novelty Search based PSO. Section 7 provides Conclusion.

\section{Brief Introduction to Novelty Search}\label{sec2}
\subsection{Basic Idea}
In 2008 Joel Lehman and Kenneth O. Stanley \cite{Leh08} proposed a new exploration algorithm which is driven by novelty of search domain. From then this search method gained quite popularity \cite{Leh10}, \cite{Leh10b},\cite{Leh11},\cite{Iz19}. The main objective of their proposed  method is to search for novel behavior in behavioral space instead of specific objective. Even in objective based problems this search method ignores the objective and look for behavioral novelty. One benefit we can observe from this algorithm is that it overcomes the obstacle of \textit{deception} and provides an \textit{open-ended evolution}. 

In any evolutionary computation, there is one or more \textit{fitness function} which measures the progress of the algorithm towards an objective. This \textit{fitness function} is also known as \textit{objective function}. Through \textit{deception} such \textit{objective functions} may prevent the objective from being searched and thus become \textit{trapped into local optima}. Objective function themselves misdirect the search into dead ends which is known as \textit{divergence}\cite{Lehke11}. \textit{Novelty Search} ignores the objective and continiously search for novel behavior as long as novelty exists in behavioral search space.
We have seen that Novelty Search asymptotically behaves like a \textit{uniform random search process in the behavior space}\cite{Don19}; but it is for sure that Novelty search is not Random Search \cite{Leh08}. Experimental results shows that Novelty Search is better than Random search.

Every swarm based algorithms or most of the evolutionary methods which are objective based faces two most critical problems - \textit{Divergence} and \textit{Convergence into local optima}. There are several approaches exist to tackle those problems \cite{Ma12}, \cite{Ka19}.Those methods not only increase the complexity of the algorithm but sometimes bring additional issues. Novelty search shows that in some problems by ignoring the \textit{objective} gives the search to achieve more explorative power and better convergence \cite{Lehke11}. 

\subsection{Novelty Search Algorithm}
How Novelty Search works is the main topic in this section. In traditional evolutionary computing the overall progress is measured by single or multi objective functions, but in novelty search instead of objective function there is a \textit{novelty metric} \cite{Lehke11} which basically measures the behavioral novelty in the search space. Novelty search is not like Exhaustive search. In most of the practical domains the novel behaviors are limited and reasonable which help novelty search to find novel behaviors in finite time rather than searching the entire search area exhaustively. 

The novelty metric measures how unique an individual\textquotesingle s behavior is. Importantly, like the fitness function, this measure must be fitted into the domain of the search space. There is an \textit{archive} which keep past behaviors of other individuals when they are highly novel. The novelty of a newly generated individual is computed with respect to the \textit{archive}.The objective is to identify how far away the new individual is from the rest of the population and its predecessors in behavior space which is the space of unique behaviors. This can be called as \textit{Sparseness} of any point into the behavior space.

One measure of \textit{Sparseness} at a point is the average distance to the k-nearest neighbors of that point, where k is a fixed parameter based on experimental problem. If the average distance to a given point in nearest neighbors is large then it is in a sparse area; if it is small then the individual is in dense area. The \textit{Sparseness} $\mu$ at any point x can be computed as follows --

\begin{equation}
\mu(x) = \frac{1}{k}\sum_{i=0}^{k} dist(x,\mu_i) 
\end{equation}

where $\mu_i$ is the ith-nearest  neighbor of x with respect to the distance metric \textit{dist}. This \textit{dist} is the behavioral difference between two individuals in the search space. During the calculation of nearest neighbor current population and the archive of novel individuals need to be taken into consideration. Solution candidates from more sparse regions of this behavioral search
space receive higher novelty scores. If novelty is sufficiently high at the location of a new individual,means above some minimal threshold value $\mu_{min}$, then the individual is entered into the permanent archive.

The current population plus the archived population jointly produce a broader sample space for current search;The gradient of search is simply towards what is new, with no other explicit objective. As we see novelty search is continuous process so we need to know when it will stop. To stop the search it is necessary to check whether each individual meets the goal criteria \cite{Lehke11}.

Novelty Search is not a Random Walk \cite{Lehke11} rather novelty search maximize novelty based on the use of \textit{archive} and \textit{novelty metric}. Novelty search gives a profound impact on the area of objective based search algorithms like Particle Swarm Optimization , Genetic Algorithm or other swarm based algorithms. If we allow particles to roam freely using novelty search and augment it with efficient yet simple objective function then we can achieve significant performance improvement by truly avoiding deception problems.

\subsection{Application Domain}

Since the introduction of Novelty search in 2008 by Lehman and Stanley \cite{Leh08} novelty search has been applied to a wide range of problems in different application domains with considerable amount of success. Those are predominantly in Robotics domain such as - single robot controller \cite{Lehke11} \cite{Mu12}, controllers for homogeneous \cite{Go13} and heterogeneous \cite{Go14} domains, robot morphologies \cite{Le14}, plastic neural networks \cite{Ri10}. A few applications outside robotics can also befound like, machine learning \cite{Na13},\cite{En13} and game content generation \cite{Li14}.

Novelty search is also applied with Particle Swarm Optimization to test the convergence, better performance and applicability on different real world domains. There are few related works in combination with Novelty search, such as  \textit{Adam Ulrich et.al} combine novelty search with particle swarm optimization  and tested CEC 2020 single objective bound-constrained optimization benchmark functions to show the effectiveness of this hybrid approach \cite{Ul21},\textit{David Martnez-Rodriguez et.al} apply PSO with Novelty search in combustion systems to find out ultra-low emissions and minimum fuel consumption \cite{Da23},\textit{Diana F. Galvao,  et. al.} combine Novelty search with PSO and apply that into Grammatical Swarm to show the performance improvement and  avoid premature convergence. They change the velocity equation with Novelty equation and instead of $P_{Best}$ , $G_{Best}$ they call those $P_{novel}$ , $G_{novel}$ \cite{Ga21}.
Novelty Search is combined with PSO in \textit{Hirad Assimi et.al's} work and applied in truss optimisation problems and experimental results show better performance over other state-of-the-art methods \cite{As21}, \textit{Clara Burgos Sim et. al} proposed an idea about applying novelty search in particle swarm optimization and tested with benchmark CEC 2005 \cite{Cl20}.

\section{Novelty Search based PSO}
\label{sec:2}
\subsection{Basic concepts}
\label{sec:21}
The proposed method has two parts. First part is Novelty Search and second part is PSO algorithm. We do not change anything on PSO algorithm. We design appropriate Novelty Search algorithm and augment it with existing PSO algorithm.

\subsection{General framework}
Our novelty search algorithm decides in which region PSO search will be applied based on novelty of that region. After that decision, PSO search will initiate. Even if PSO finds better results than in previous iteration, search will not stop until no novel area is left.

Let\textquotesingle s explain our approach step by step:
First, we decide how many particle will be deployed initially on the application search area. Suppose, we deployed 5 particles on the application search region e.g. $P_1,P_2,P_3,P_4,P_5$. We call them \textit{\textbf{Leader Particles}}. These five \textit{Leader particles} will initiate PSO search once they find novel area for searching. Movement of \textit{Leader particles} solely based on novelty search. 

After randomly distributed over search area, each Leader particle $P_i$ will check novelty of their \textit{search region} with other particle\textquotesingle s \textit{search region}. By \textit{search region}, we mean to say that a circular area whose center is the Leader particle\textquotesingle s position and radius r is pre-determined by the application. The concept of $P_i$'s \textit{search region} is better explained with a diagram shown in figure-1.

\begin{figure}[h]
	\centering
	\begin{tikzpicture}
	\begin{scope}
	\draw[blue]node[black]{$P_i$}(0,0) circle (2.0cm);
	\draw[orange,thick,->](0.2,0) -- node[black,above]{r}(2.0,0);
	\end{scope}
	\end{tikzpicture}
	\caption{Circular region defined by Leader particle $P_i$ with radius r.}
\end{figure}
\begin{figure}[h]
	\centering
	\begin{tikzpicture}
	\begin{scope}
	\draw[blue](1,1)node[black]{$P_1$} circle (1.0cm);
	\draw[orange,thick,->](1.2,1) -- node[black,above]{r}(2.0,1);
	
	\draw[green](-2,-2)node[black]{$P_2$} circle (1.0cm);
	\draw[orange,thick,->](-1.7,-2) -- node[black,above]{r}(-1.0,-2);
	
	\draw[black](3,-2)node[black]{$P_3$} circle (1.0cm);
	\draw[orange,thick,->](3.2,-2) -- node[black,above]{r}(4.0,-2.0);
	
	\draw[magenta](1.6,-2)node[black]{$P_4$} circle (1.0cm);
	\draw[orange,thick,->](1.8,-2) -- node[black,above]{r}(2.6,-2.0);
	
	\draw[red](-2,3)node[black]{$P_5$} circle (1.0cm);
	\draw[orange,thick,->](-1.8,3) -- node[black,above]{r}(-1.0,3);
	\draw[->,thick] (-4,0)--(4,0) node[right]{$x$};
	\draw[->,thick] (0,-4)--(0,4) node[above]{$y$};
	\end{scope}
	\end{tikzpicture}
	\caption{Randomly distributed five Leader particles $P_1,P_2,P_3,P_4,P_5$ with radius r in search domain.}
\end{figure}

In Figure-2, We randomly distribute five Leader particles $P_1$ to $P_5$ in the two dimensional application search domain. Here, we need to mention that our experimental optimization test functions are not two dimensional and most of the times they are either 10,30,50 or more than that. That means it is not always circle(two dimensional) or sphere (three dimensional) rather it is \textit{N-sphere} or \textit{Hypersphere} in N+1-Dimensional Euclidean space.

For N+1 Euclidean dimensional Space, if the points are represented as ($x_1,x_2,x_3,...,x_{N+1}$) then the \textit{n-space}, $S^N(r)$, is represented by the equation --

\begin{equation}
r^2 = \sum_{i=1}^{N+1} (x_i - c_i)^2 
\end{equation}

where, c = ($c_1,c_2,...c_{N+1}$) is a center point, and r is the radius.

For mathematical representation we mention equation-(2). But for our experiment we only need the calculation to check weather any N-dimensional point lies inside, outside or on the boundary of the \textit{Hypersphere}, which is not different from calculating the same, with circle. So for simplicity we keep our discussion in the concept of a circle(in two dimensional problem space) as per figure-(2).

Now,each $P_i$ defines its region based on radius r. Each $P_i$ will initiate PSO search based on its \textit{novelty score}.  \textit{Novelty score} and all other related terms are discussed in  \textit{Novelty of a Search Region} as discussed in section(section 3.3).	.

Once a Leader particle $P_i$ finds that, its search region is novel in comparison with other $P_j$ \textquotesingle s (here, j$\ne$i and j = 1 to n) search region, that leader $P_i$ will initiate PSO algorithm to find better solution of the given optimization problem.

What PSO algorithm will be used and how they will perform, will be discussed in section 3.3.4 in detail.

If any Leader particle $P_i$ finds that, its search region is not novel enough compare to any other $P_j$\textquotesingle s (Where, j$\ne$i and j = 1 to n) search region then it will not perform any PSO search; rather it will re-compute its position.

Let us shed some light on what does it mean by \textit{not novel enough}. It means that search region is either already searched by other \textit{Leader particles} or this \textit{Leader particle} in some previous iteration. Repeatedly searching the same region instead of keeping open some other unsearched area is not the purpose of novelty search. Here, one may argue that, many PSO instances are roaming around; as there is no modification in PSO then how it eliminates premature convergence? One thing we need to make clear here, novelty search helps to jump from one place to another based on their behavioral novelty, and the premature convergence happens because of lack of diversity, which is not happening during novelty search. Rather we can ask the question how PSO will reach into the optimal solution, what happens if it gets stuck in local optima? It will be clear more when we will discuss different categories of novelty in section 3.3.2.    

The major difference here is that, if any leader particle finds better solution than its earlier iteration the search will not stop there; and the \textit{better solution} achieved so far will not communicate with other leader particles immediately; rather \textit{Leader particles} store their solution in an archive and jumps to the next novel location and keep on searching until there are no novel search region left. By this approach we can explore more and more unknown territory and will never be stuck in any local optimum valley. 

Does it looks like \textit{Random Restart} of a group of multiple PSO? Lets discuss the major difference between Random Restart plus some local optimizer and our approach. In our algorithm a group PSO is used under the name \textit{Leader particles}. Those \textit{Leader particles} control their corresponding PSO groups. They decide where to search based on novelty. Novelty search is a divergence algorithm.  Novelty search helps our algorithm to cover the  unknown territory much quicker as it is driven by difference between two locations. The more difference means more suitable for performing the PSO search. Not novel  means no unnecessary search  will happen. But in case of random restart it will not guarantee that same search region will not reappear again. Not only that our purpose for novelty search is exploration. No number of random restart can assure the coverage of entire search area no matter how efficient local optimizer we use.

In PSO, stuck into local optimum for a considerable period of time is the major concern; especially for those optimization function where there are many more local valleys and global solution is far from local solution. As PSO is objective based search method, any particle finds any so-far-best solution other particles will automatically jump towards that location without exploring other areas. Novelty search overcomes this problem as ''\textit{objective-free}'' searching is its inherent nature. No matter what PSO search will result, Novelty search forces leader particles to keep on searching for any better results. In this way the present approach becomes better explorative algorithm than existing PSO methods. 

In summary the present algorithm works as follows;\\
\textbf{Step 1}: Deployment of \textit{''Leader particles''} $P_i$ into the application search area.\\
\textbf{Step 2}: Based on application based radius value ( r ) a circular region is defined by leader particle $P_i$, and $P_i$ is positioned at the center of the search region.\\
\textbf{Step 3}: Each leader particle $P_i$ will compute its \textit{novelty score} with other leader particles. 	\\
\textbf{Step 4}: if $P_i$ \textquotesingle s search region is novel compare to other particle\textquotesingle s search region it will perform PSO search on that search region. A global archive \textit{NS\_archive} for storing locations is used. \textit{NS\_archive} is a multidimensional global vector.Each $P_i$ stores its location  into \textit{NS\_archive} whenever it is performing its PSO search. \\
\textbf{Step 5}: if $P_i$ \textquotesingle s search region is NOT novel compare to other particle?s search region, $P_i$ will re-compute new position without initiating any PSO search.\\
\textbf{Step 6}:  This process will continue until no ''\textit{Novel}'' search area is left in the application search region.\\

\subsection{Novelty of a Search Region}
In this section we will discuss how novelty of a region which is identified and controlled by leader particle $P_i$ is computed.
\subsubsection{Novelty Computation Procedure}
First let us clarify why we compute novelty of any region. In simple word Novel region means that  region which is  still not searched by any \textit{PSO team}. Here by \textit{PSO team} means , a group of particles whose leader is Leader Particle $P_i$. Once   $P_i$ finds that its current region is not previously searched by any leader particle  and none of the other leader particles currently holding that region , $P_i$ will initiates PSO searching. We augment novelty search with pso search because novelty search gives PSO more explorative power. To search unknown territory we need an algorithm which is not governed by any objective based method. Novelty search is perfectly fit for that. It is objective free and its simplistic approach helps PSO to search more diverse area.

Let's discuss in detail how we compute novelty. After leader particles are distributed over the application search area, they first compute its region. Here region means the circular area whose center is leader particle's (i.e. $P_i$'s) coordinate position and radius (r) of that circular area. Radius r is obtained from the need of the application domain. In our experiment we use a dynamic approach for computing r. We first use r = 1, a unit value then increase this value by 0.1 as per our need. Larger r value indicates that PSO search takes longer time and lesser r value means novelty search takes more time to cover the whole search space. So we try to keep this value moderate. We will calculate the impact of r-value and find out what will be the best possible value of r for the algorithm to give best result of our work.  

By referring figure-(1), we see that the center of the circular area hold by particle $P_1$ is $P_1$'s coordinate position and r is its radius. Region is the area - $\pi r^2$. We calculate this area for the purpose that when leader particle initiates PSO search , it randomly distributes particles inside that area. We need to check that no particle is outside of that area.  For that we use equation- (4), which is the distance d between two particles. If $d_i > $ r then the particle is outside and we reject the particle else we  accept the particle.

When every leader particle knows its region it computes its novelty with respect to other leader particle. Novelty of a leader particle is judged by its novelty score as stated in section 3.3.2.

We explain the novelty computation as follows; We consider the example of 5 leader particles of figure-(2). Suppose first leader particle $P_1$ calculates its novelty. For that,  $P_1$ first decides whether $P_1$ has already searched that area or any other leader particle searched that area earlier by calculating novelty score for each of them. We call novelty calculation between two particles position as \textit{\textbf{"novelty score(ns)"}}. $P_1$ calculates novelty score(ns) for its previous position and also for other particle's previous positions one by one. This ns calculation is basically the distance calculation according to the equation-(4). Once $P_1$ gets all the ns value from $P_1$ to previous position of  $P_2,P_3,P_4,P_5$ and its own previous position, $P_1$ will compute \textit{average ns value} according to the equation-(5). All the previous locations where successful PSO search are already occurred are stored in \textit{NS\_archive}. So each $P_i$, after deployment into the search domain, consult its own position with the locations stored in  \textit{NS\_archive}.
If the novelty score(ns) is below \textit{\textbf{"novelty\_threshold"}} value, it indicates that $P_1$\textquotesingle s region is already searched. Therefore there is no point in searching the same region again. So $P_1$ stops calculating novelty score any further and sends \textbf{\textit{recompute message to itself}} as shown in the figure-(3). Otherwise no action is taken and $P_1$ moves to next step. The \textit{novelty\_threshold} is discussed in section 3.3.3. Let us consider that it is a numerical value with which comparison of \textit{ns} happens.

In figure-(3), all dashed circles are leader particle's previous position.We renamed them as $P_1',P_2',P_3','P_4',P_5'$. Current position of $P_1$ is same as $P_1$. Consider that $ns_1,ns_2,ns_3,ns_4$ and $ns_5$ are novelty score of $P_1$  for the leader particles $P_1,P_2,P_3,P_4$ and $P_5$'s previous position respectively. Suppose $ns_1$ is below novelty threshold value $Th_N$ then $P_1$ sends \textit{recompute} message to itself as shown in the figure-(3). This is indicated by a violet colored self loop marked in $P_1$. $recompute(P_1')$ indicates that $P_1$ sends a \textit{recompute} message to itself because $P_1$ searched this place previously.

\begin{figure}[h]
	\centering
	\begin{tikzpicture}
	\begin{scope}
	\draw[blue](3,2)node[black]{$P_1$} circle (1.0cm);
	\draw[orange,thick,->](3.2,2) -- node[black,below]{r}(4.0,2);
	\draw[blue, thick,->](3.2,2) -- node[black,above]{$ns_1$}(1.3,1);
	\draw[green, thick,->](3.2,2) -- node[black,below]{$ns_2$}(-1.7,-2);
	\draw[black, thick,->](3.2,2) -- node[black,above]{$ns_3$}(3.2,-2);
	\draw[magenta, thick,->](3.2,2) -- node[black,below]{$ns_4$}(1.8,-2);
	\draw[red, thick,->](3.2,2) -- node[black,below]{$ns_5$}(-1.8,3);
	
	\draw [->,ultra thick, violet](3,3) to [out=430,in=100,looseness=8] node[above left, black]{$recompute(P_1')$}(2.4,2.8);
	
	\draw[blue,dashed](1,1)node[black]{$P_1'$} circle (1.0cm);
	\draw[orange,thick,->](1.3,1) -- node[black,above]{r}(2.0,1);

	\draw[green,dashed](-2,-2)node[black]{$P_2'$} circle (1.0cm);
	\draw[orange,thick,->](-1.7,-2) -- node[black,above]{r}(-1.0,-2);
	
	\draw[black,dashed](3,-2)node[black]{$P_3'$} circle (1.0cm);
	\draw[orange,thick,->](3.2,-2) -- node[black,above]{r}(4.0,-2.0);
	
	\draw[magenta,dashed](1.6,-2)node[black]{$P_4'$} circle (1.0cm);
	\draw[orange,thick,->](1.8,-2) -- node[black,above]{r}(2.6,-2.0);
	
	\draw[red,dashed](-2,3)node[black]{$P_5'$} circle (1.0cm);
	\draw[orange,thick,->](-1.8,3) -- node[black,above]{r}(-1.0,3);
	
	\draw[->,thick] (-4,0)--(4,0) node[right]{$x$};
	\draw[->,thick] (0,-4)--(0,4) node[above]{$y$};
	\end{scope}
	\end{tikzpicture}
	\caption{Novelty Score checking with the leader particle \textquotesingle s previous positions.}
\end{figure}

When $P_1$ finds that no particle, not even itself search that area previously, then it checks whether anyone among $P_2, P_3,P_4$ and $P_5$ are currently holding the same region. To check that, $P_1$ computes novelty score with $P_2,P_3,P_4$ and $P_5$ one by one. If any novelty score is below \textit{"novelty\_threshold"} value then $P_1$ sends a \textbf{\textit{"recompute" message to itself}} as shown in figure-(4), but this time recomputation is done due to closely overlapping position with other particles.

\begin{figure}[h]
	\centering
	\begin{tikzpicture}
	\begin{scope}
	\draw[blue](3,2)node[black]{$P_1$} circle (1.0cm);
	\draw[orange,thick,->](3.2,2) -- node[black,below]{r}(4.0,2);
	\draw[green, thick,->](3.2,2) -- node[black,below left]{$ns_2$}(2.2,-3);
	\draw[black, thick,->](3.2,2) -- node[black,above right]{$ns_3$}(3.2,-2);
	\draw[magenta, thick,->](3.2,2) -- node[black,above left]{$ns_4$}(2.7,1.5);
	\draw[red, thick,->](3.2,2) -- node[black,below]{$ns_5$}(-2.0,3.5);
	
	\draw [->,ultra thick, violet](3,3) to [out=430,in=100,looseness=8] node[above left, black]{$recompute(P_1')$}(2.4,2.8);
	

	\draw[green](2,-3)node[black]{$P_2$} circle (1.0cm);
	\draw[orange,thick,->](2.2,-3) -- node[black,above]{r}(3.0,-3);
	
	\draw[black](3,-2)node[black]{$P_3$} circle (1.0cm);
	\draw[orange,thick,->](3.2,-2) -- node[black,above]{r}(4.0,-2.0);
	
	\draw[magenta](2.5,1.5)node[black]{$P_4$} circle (1.0cm);
	\draw[orange,thick,->](2.7,1.5) -- node[black,above]{r}(3.5,1.5);
	
	\draw[red](-2.2,3.5)node[black]{$P_5$} circle (1.0cm);
	\draw[orange,thick,->](-2.0,3.5) -- node[black,above]{r}(-1.2,3.5);
	
	\draw[->,thick] (-4,0)--(4,0) node[right]{$x$};
	\draw[->,thick] (0,-4)--(0,4) node[above]{$y$};
	\end{scope}
	\end{tikzpicture}
	\caption{Novelty Score checking with other leader particle \textquotesingle s. \textit{recompute} message is sent to itself.}
\end{figure}

In the figure -(4), $P_1$ to $P_5$ all leader particle's position are updated. Now they are in different positions than figure-(2). $P_1$ computes novelty score with all the leader particles one by one, as we can see $ns_1,ns_2,ns_3,ns_4$ and $ns_5$. But $ns_4$ shows that particle $P_4$ is too close to particle $P_1$. Their searching region is largely overlapping. As searching the almost same region by two leader particles is not meaningful, $P_1$ sends a \textit{recompute} message to itself. we can see a violet colored arrow from $P_1$ to $P_1$ indicating the re-computation of position.  

Here, two scenario can happen during ns calculation. one is - sequential , that means first $P_1$ calculates its ns then $P_2$, $P_3$,$P_4$, and $P_5$. If that happen then $P_1$ only need to compute ns with the previous positions of all the leader particles including itself by consulting \textit{NS\_archive} table. No need to compute ns with current positions of all the leader particles. Second one is -  parallel, if the environment allows all the particles parallel computation then $P_1$ will check \textit{NS\_archive} table for the current position of all other leader particles during its ns calculation.  

Once all the novelty score is calculated, leader particle calculates average of all the novelty score. As per our above figure particle $P_1$ calculates  average of $ns_2,ns_3,ns_4$ and $ns_5$ as follows.

\begin{equation}
N_i = \sum_{j=1}^{n} \frac{ns_j}{n-1} 
\end{equation}
where $j \neq i$ and n is the number of leader particles.

Now this average of novelty score($N_i$) is called \textbf{novelty of $P_i$}.

The average calculation of all novelty score gives us the indication weather $P_i$ will search that region or not. After calculating the novelty value, $P_i$ will check whether this $N_i$ is above the \textit{novelty\_threshold($Th_N$)} value. If $N_i$ $\geq$ $Th_N$ then the region holds by leader particle $P_i$ is novel enough to initiate PSO search else $P_i$ will again recompute its position.

\subsubsection{Novelty score calculation}
Novelty score, calculated between two particles, indicates how close two regions are. Basically it is the 
\textit{Euclidean distance between two leader particles}. If two leader particles are too close that means two regions are mostly overlap. Two highly overlapped regions means two PSO teams performing their search operation on that region where most of the area is common. Searching the same area by two PSO teams wastes the searching capability of swarm particles of both the teams. If leader particles $P_i$  are in enough distance it means they are enough novel and searching the unsearched area is quite justified.
The formula for Euclidean distance between two leader particles in a multidimensional environment is as follows - 
\begin{equation}
d_i^2 = \sum_{v=1}^{dim} (X_{vi} - X_{vj})^2 
\end{equation}
where, $d_i$ is the distance between particle $P_i$ and $P_j$. $v = 1$ to $dim$ is the dimension which depends on application. $X_{vi}$ and $X_{vj}$ is the coordinate positions of particle  $P_i$ and $P_j$.

In figure-(5), $d_1$ is the distance between $P_1$ and $P_2$. Distance is calculated based on two dimensional search space.

\begin{figure}[h]
	\centering
	\begin{tikzpicture}
	\begin{scope}
	\draw[blue,fill=LimeGreen!10](3,2)node[black]{$P_1$} circle (1.5cm);
	\draw[orange,thick,->](3.2,2) -- node[black,below right]{r}(2,0.9);
	
	\draw[green,fill=Apricot!10](-2,2 )node[black]{$P_2$} circle (1.5cm);
	\draw[orange,thick,->](-2.2,2) -- node[black,below left]{r}(-1.0,0.9);
	
	\draw[violet,thick,<->](3.2,2) -- node[black,below right]{$d_1$}(-2.2,2);
	
	\node [] at (0.8,-1.5){$d_1 = \sqrt{(X_2 -X_1)^2 - (Y_2 -Y_1)^2}$};
	\draw[->,thick] (-4,0)--(4,0) node[right]{$x$};
	\draw[->,thick] (0,-4)--(0,4) node[above]{$y$};
	\end{scope}
	\end{tikzpicture}
	\caption{Distance between two regions $P_1$ and $P_2$ in a 2 dimensional search space.$d_1 = \sqrt{(X_2 -X_1)^2 - (Y_2 -Y_1)^2}$ }
\end{figure}
As distance clearly influence novelty, so how much closeness is novel and how much is not is a matter of discussion. There are several stages of novelty. 

Let us discuss novelty score scenario one by one.

\begin{itemize}
	\item \textbf{Case 1}: [ d $\geq$ 2*r ] : When distance(d) between two regions is grater than or equal to twice of radius(r) of the region then this two region is perfectly distinct, highly novel and good candidates for PSO searching. So both $P_1$ and $P_2$ will initiate PSO searching. Figure -(6) showing this Case 1.
	
	\begin{figure}[h]
		\centering
		\begin{tikzpicture}
		\begin{scope}
		\draw[blue,fill=LimeGreen!10](1.5,1.5)node[black]{$P_1$} circle (1.5cm);
		\draw[VioletRed,fill=Apricot!10](-1.5,1.5 )node[black]{$P_2$} circle (1.5cm);
		
		\draw[orange,thick,->](1.4,1.5) -- node[black,below right]{r}(0.45,0.45);
		\draw[orange,thick,->](-1.4,1.5) -- node[black,below left]{r}(-0.45,0.45);
		
		\draw[Mulberry,thick,<->](-1.4,1.5) -- node[black,above right]{$d_1$}(1.4,1.5);
		\draw[BlueGreen,thick,<->](-1.3,1.3) -- (1.3,1.3) node[black,fill = white,midway]{$2*r$};
		
		\node [draw] at (-1.0,3.5){Case 1: When $d_1 \geq 2*r$};
		
		\draw[->,thick] (-4,0)--(4,0) node[right]{$x$};
		\draw[->,thick] (0,-4)--(0,4) node[above]{$y$};
		
		\node [draw] at (0,-1.0){Completely Novel area for $P_1$ and $P_2$};
		\draw[->,thick] (0.1,-0.8)--(1.5,0.8) node[]{};
		\draw[->,thick] (0.1,-0.8)--(-1.5,0.8) node[]{};

		\end{scope}
		\end{tikzpicture}
		\caption{Two regions at 2*r distance apart are perfect candidates for PSO searching.}
	\end{figure}
	
	\item \textbf{Case 2}: [ r $\le$ d $\le$ 2*r ] : If the distance d is higher than the radius(r) but lower than the 2*r then though there is some common area for searching but still a lot of area is different and needs searching. The exact scenario is shown in figure-(7).
	
	\begin{figure}[h]
		\centering
		\begin{tikzpicture}
		\draw[blue,fill=LimeGreen!10](1.0,1.5)node[black]{$P_1$} circle (1.5cm);
		\draw[VioletRed,fill=Apricot!10](-1.5,1.5 )node[black]{$P_2$} circle (1.5cm);
		
		\draw[orange,thick,->](1.1,1.5) -- node[black,below right]{r}(0.28,0.28);
		\draw[orange,thick,->](-1.4,1.5) -- node[black,below left]{r}(-0.45,0.45);
		
		\node [draw] at (-1.0,3.5){Case 2: When r $\le$ $d_1$ $<$ 2*r };
		
		\begin{scope}
		\clip (1.0,1.5) circle (1.5cm);
		\fill[color=RoyalPurple!50] (-1.5,1.5) circle (1.5cm);
		\end{scope}
		
		\draw[Mulberry,thick,<->](-1.3,1.5) -- node[black,fill = white,midway]{$d_1$}(0.9,1.5);
		\draw[BlueGreen,thick,<->](-1.5,1.1) -- (1.1,1.1) node[black,fill = white,midway]{$< 2*r$};
		
		\draw[->,thick] (-4,0)--(4,0) node[right]{$x$};
		\draw[->,thick] (0,-4)--(0,4) node[above]{$y$};
		
		\node [draw] at (0,-1.8){$P_1 \cap P_2 < 50\%$};
		\draw[->,thick] (0.1,-1.6)--(-0.3,0.4) node[]{};
		
		\node [draw] at (-2.5,-1.0){Novel area for $P_2$ $> 50\%$};
		\draw[->,thick] (-2.6,-0.8)--(-2.3,0.6) node[]{};
		
		\node [draw] at (2.5,-1.0){Novel area for $P_1$ $> 50\%$};
		\draw[->,thick] (2.6,-0.8)--(2.1,0.8) node[]{};
		
		\end{tikzpicture}
		\caption{Two regions are overlapped. This less than $50\%$ Common area is redundant region for searching.}
	\end{figure}
	
	\item \textbf{Case 3}: [ d = r ] : If the distance d is exactly same as radius(r) of the region, that means $50\%$ search region is matched. So we are left with other $50\%$ which is unique and need to be searched. Now it is upto the application whether we will continue to search that area or not. In general thinking, it will be good if we initiate PSO search on that $50\%$ search region. This  $50\%$ novelty condition  is a good condition for PSO search. This scenario is shown in figure-(8).
	
	\begin{figure}[h]
		\centering
		\begin{tikzpicture}
		
		\draw[blue,fill=LimeGreen!10](0.75,1.5)node[black]{$P_1$} circle (1.5cm);
		
		\draw[VioletRed,fill=Apricot!10](-0.75,1.5 )node[black]{$P_2$} circle (1.5cm);

		\node [draw] at (-1.0,3.5){Case 3: When $d_1 = r$ };
		
		\begin{scope}
		\clip (0.75,1.5) circle (1.5cm);
		\fill[color=RoyalPurple!30] (-0.75,1.5) circle (1.5cm);
		\end{scope}
		
		\draw[orange,thick,->](0.76,1.5) -- node[black,below right]{r}(2.1,2.3);
		\draw[orange,thick,->](-0.76,1.5) -- node[black,below left]{r}(-2.2,1.5);
		
		\draw[Mulberry,thick,<->](-0.74,1.5) -- node[black,above right]{$d_1$}(0.74,1.5);
		\draw[BlueGreen,thick,<->](-0.74,1.3) -- (0.74,1.3) node[black,fill = white,midway]{$r$};
		
		\draw[->,thick] (-4,0)--(4,0) node[right]{$x$};
		\draw[->,thick] (0,-4)--(0,4) node[above]{$y$};
		
		\node [draw] at (0,-1.8){$P_1 \cap P_2 = 50\%$};
		\draw[->,thick] (-0.1,-1.6)--(-0.3,0.4) node[]{};
		
		\node [draw] at (-2.5,-1.0){Novel area for $P_2$ $= 50\%$};
		\draw[->,thick] (-2.4,-0.8)--(-1.7,0.6) node[]{};
		
		\node [draw] at (2.5,-1.0){Novel area for $P_1$ $= 50\%$};
		\draw[->,thick] (2.6,-0.8)--(1.9,0.8) node[]{};
		
		\end{tikzpicture}
		\caption{Two regions of $P_1$ and $P_2$ are overlapped exactly $50\%$.  So both $P_1$ and $P_2$ left with $50\%$ unsearched unique area.}
	\end{figure}
	
	\item \textbf{Case 4}: [ 0 $ <$ d $<$ r ] : If the distance d is less than the radius(r) of the region, that means  more than $50\%$ search region is matched between two leader particles. So we are left with  very less amount of search region which is unique. As the novel region is very less so either we recompute a new position by leaving this region unsearched or we keep on performing PSO search on this area. Here \textit{novelty\_threshold}($Th_N$) is required, which will determine what we will do. This case is shown in figure-(9).
	
	\begin{figure}[h]
		\centering
		\begin{tikzpicture}
		
		\draw[blue,fill=LimeGreen!10](0.25,1.5)node[black]{$P_1$} circle (1.5cm);
		
		\draw[VioletRed,fill=Apricot!10](-0.25,1.5 )node[black]{$P_2$} circle (1.5cm);

		\node [draw] at (-1.0,3.5){Case 4: When 0 $<$ d $<$ r };
		
		\begin{scope}
		\clip (0.25,1.5) circle (1.5cm);
		\fill[color=RoyalPurple!30] (-0.25,1.5) circle (1.5cm);
		\end{scope}
		
		\draw[orange,thick,->](0.26,1.5) -- node[black,above right]{r}(1.76,1.5);
		\draw[orange,thick,->](-0.26,1.5) -- node[black,above left]{r}(-1.76,1.5);
		
		\draw[Mulberry,thick,<->](-0.24,1.5) -- node[black,above right]{$d_1$}(0.24,1.5);
		\draw[BlueGreen,thick,<->](-0.26,1.3) -- (-1.76,1.3) node[black,fill = white,midway]{$r_2$};
		\draw[BlueGreen,thick,<->](0.26,1.3) -- (1.76,1.3) node[black,fill = white,midway]{$r_1$};
		
		\draw[->,thick] (-4,0)--(4,0) node[right]{$x$};
		\draw[->,thick] (0,-4)--(0,4) node[above]{$y$};
		
		\node [draw] at (0,-1.8){$P_1 \cap P_2$ $\gg$ $50\%$ and $d_1$ is $\ll$ $r_1$ and $r_2$};
		\draw[->,thick] (-0.1,-1.6)--(-0.3,0.4) node[]{};
		
		\node [draw] at (-2.5,-1.0){Novel area for $P_2$  $\ll$ $50\%$};
		\draw[->,thick] (-2.4,-0.8)--(-1.4,0.6) node[]{};
		
		\node [draw] at (2.5,-1.0){Novel area for $P_1$ $\ll$ $50\%$};
		\draw[->,thick] (2.6,-0.8)--(1.5,0.8) node[]{};
		
		\end{tikzpicture}
		\caption{Two regions of $P_1$ and $P_2$ are overlapped more than $50\%$.  So both $P_1$ and $P_2$ left with very less unsearched unique area.}
	\end{figure}
	
	\item \textbf{Case 5}: [ d = 0 ] : If the region holds by the leader particles completely overlap with each other then there will be no novel region left. So no need of initiating PSO search in that region.Basically one of them will search that area and other leader particle will move to other region by re-computing its position. Complete overlapping is very rare thing to occur, because we need to remember that positions are distributed randomly in the application search area. So it is very much unlikely that two random numbers are same. So this case which is shown in figure-(10) is very uncommon incident.
	
	\begin{figure}[h]
		\centering
		\begin{tikzpicture}
		
		\draw[blue,fill=LimeGreen!10](0.0,1.5)node[black]{$P_1$} circle (1.5cm);
		
		\draw[VioletRed,fill=Apricot!10](0.0,1.5 )node[black]{$P_2$} circle (1.5cm);

		\node [draw] at (-1.0,3.5){Case 5: When d = 0 };
		
		\begin{scope}
		\clip (0.0,1.5) circle (1.5cm);
		\fill[color=RoyalPurple!30] (0.0,1.5) circle (1.5cm);
		\end{scope}
		
		\draw[orange,thick,->](0.0,1.5) -- node[black,below right]{r}(1.5,1.5);
		\draw[orange,thick,->](0.0,1.5) -- node[black,below left]{r}(-1.5,1.5);
		
		\draw[Mulberry,thick,<->](0,1.5) -- node[black,above right]{$d_1 = 0$}(0,1.5);
		\draw[BlueGreen,thick,<->](-1.5,1.2) -- (1.5,1.2) node[black,fill = white,midway]{$2*r$};
		
		\draw[->,thick] (-4,0)--(4,0) node[right]{$x$};
		\draw[->,thick] (0,-4)--(0,4) node[above]{$y$};
		
		\node [draw] at (0,-1.8){$P_1 \cap P_2$ = $100\%$};
		\draw[->,thick] (-0.1,-1.6)--(-0.3,0.4) node[]{};
		
		\node [draw] at (0,-3.0){No Novel area left for $P_1$ and $P_2$!};
		\end{tikzpicture}
		\caption{Two regions of $P_1$ and $P_2$ are completely overlapped. So both $P_1$ and $P_2$ left with zero unsearched unique area.}
	\end{figure}

\end{itemize}

By observing all possible novelty cases, we can represent them in a tabular form as shown in Table-1.

\begin{table*}[ht]
	\caption{Tabular representation of different novelty cases and their categories.}
	\centering
	\begin{tabular}{ ||c|c|c|c|c|| }
		\hline
		\textbf{Case Number} & \textbf{Condition} & \textbf{Novelty Percentage} & \textbf{Category} & \textbf{Remark} \\
		\hline\hline
		\textbf{1} & d $\ge$ 2*r & 100\% & \colorbox{green}{Highly Novel} & Perfect for PSO search\\
		\hline
		\textbf{2} & r $<$ d $<$ 2*r & 51 - 99\% & \colorbox{green}{Moderately Novel} & Good for PSO search\\
		\hline
		\textbf{3} & d = r & 50\% & \colorbox{green!30}{Typically Novel} & OK for PSO search\\
		\hline
		\textbf{4} & 0 $<$ d $<$ 2*r & 1 - 49\% & \colorbox{yellow!30}{Low Novel} & Less Likely for PSO search\\
		\hline
		\textbf{5} & d = 0 & 0\% & \colorbox{yellow}{Not Novel} & Not for PSO search\\
		\hline
		
	\end{tabular}
	
\end{table*}	

After careful consideration of all possible novelty cases, we come to a formula for calculating average novelty score for particle $P_i$ that is $ns_i$ is as follows -

\begin{equation}
ns_i = (\frac{d_i}{2*r}) *100%
\end{equation}

Now we come to the point where we need novelty threshold value. It is totally based on application. Clearly from the table -1, if we set very low value of novelty threshold like case -4 (1-49\%) that means we allow two PSO teams to search closely. If our application demand closeness we can set low novelty threshold value. If two regions are highly novel  like case-1(100\%) as per table-1, that signifies that these two regions are not overlapping and both the leader particle can initiate searching.

So initially, having moderate threshold value is good for exploration, when there is vast area unsearched. And if we keep decreasing novelty threshold value as time passes more rigorous search will happen which will include boundary areas of region. If we set $Th_N$ is high(closely 100\%) then the following scenario like the figure-(11) will happen. We can see the triangle shaped area is left unsearched. Like this many more fragments of search area will be left unsearched. The problem of setting high $Th_N$ value creates many small unsearched fragments that is why decreasing the threshold improves the situation and left no unsearched  area also makes our algorithm more diverse.

One important point here is when we stop novelty search? In our experiment we keep on performing novelty search until all the Leader particles' novelty score according to the equation-(5)  is in "Low Novel" category as per table -1. This gives us excellent results in more complex multidimensional and multi-modal functions.

\begin{figure}[h]
	\centering
	\begin{tikzpicture}
	
	\draw[blue,fill=LimeGreen!10](-1.5,1.5)node[black]{$P_1$} circle (1.5cm);
	
	\draw[VioletRed,fill=Apricot!10](1.5,1.5 )node[black]{$P_2$} circle (1.5cm);
	
	\draw[Sepia,fill=Cyan!10](0,-1.1)node[black]{$P_3$} circle (1.5cm);

	\draw[->,thick] (-4,0)--(4,0) node[right]{$x$};
	\draw[->,thick] (0,-4)--(0,4) node[above]{$y$};

	\node [draw] at (-1.5,-3){Triangle shaped unsearched region};
	\draw[->,thick] (-1.4,-2.8)--(-0.2,0.6) node[]{};
	
	\end{tikzpicture}
	\caption{Two regions of $P_1$ and $P_2$ are completely overlapped. So both $P_1$ and $P_2$ left with zero unsearched unique area.}
\end{figure}

\subsubsection{Novelty Threshold}

This \textit{novelty\_threshold} or $Th_N$ value is pre-determined by the application. A simple idea behind  \textit{novelty\_threshold} is how much close two PSO teams will search? There is no point of searching the same place by two different PSO teams. So obviously one PSO team will be withdrawn from that region and replaced to some other search area. If $Th_N$ value is very low then two PSO team will search very closely and almost same place, and if the $Th_N$ value is very high then two PSO team will be far away and lots of search area will be unsearched. So we will keep this threshold value in a moderate level so that entire area will be searched properly  and no area is left unsearched.

\subsubsection{Variant of Particle Swarm Optimization used in our algorithm}

We can use any variant of particle swarm optimization algorithm for our method. Every PSO algorithm has two parts; one \textit{exploration} and another \textit{exploitation}. Exploration helps pso to search more diverse area where as exploitation helps to go deeper if any promising search area found. According to the latest benchmark test optimization functions \cite{lia13} most of the PSO variants suffers when the test function becomes complex, non-seperable and asymmetrical. The latest standard for PSO algorithm which is known as "SPSO11"\cite{Cle11}, also fails to deliver promising performance on those functions where there is huge number of local minima and second global minimum is far from first true optimum \cite{Zam13}.

Choosing a variant which is good in terms of "Exploitation" is a perfect fit for our method, because novelty search do the "Exploration" part far better than any other algorithm. Once Novelty search finds out a location for performing PSO search, any good exploitive PSO will do the job.

In our case we took BareBones-PSO variant developed by Kennedy \cite{Ken03}. The reason behind this is its simplicity. This method is Velocity free, and its particles positions are based on Gaussian distribution. Its simplified position equation is as follows -
\begin{equation}
\small x_{ij}(t+1) \sim N\biggl(\frac{pbest_{ij}(t) + gbest_j(t)}{2},|pbest_{ij}(t) - gbest_j(t)|\biggr)
\end{equation}

where \\
\begin{align}
\rho = |pbest_{ij}(t) - gbest_j(t)| \nonumber \\ \sigma = \frac{pbest_{ij}(t) + gbest_j(t)}{2}
\end{align}

$\rho$ is the mean of $pbest_{ij}(t)$ and $gbest_j(t)$ and $\sigma $ is standard deviation of $pbest_{ij}(t)$ and $gbest_j(t)$ and N( $\sigma,\rho$ ) is the Gaussian Distribution with $\sigma$ and $\rho$. $pbest_{ij}(t)$ and $gbest_j(t)$ are as usual pso terms of local best position and global best position at time t. 

To discuss more on other aspects of  Bare Bones PSO like maximum iterations and number of restarts etc is not the objective of our paper. Please follow \cite{Ken03}, \cite{Pan08}, \cite{Tim11} for more information about Bare Bones PSO and its Theoretical analysis.

After Novelty Search, when leader particle calls Bare-Bones PSO, it is upto Bare-Bones PSO which will generate a group of random particles on that region controlled by leader particle. If the search region is comparatively small than other PSO variants then its search becomes faster and deeper.

After the completion of search if global best position is below fitness threshold value ($Th_f$) then that position will be kept in record, that is in the \textit{PSO\_archive}. This \textit{PSO\_archive} gives us the optimum value achieved by our algorithm and it is a global table. Initially it looks that this \textit{PSO\_archive} is huge table with million of values as the algorithm runs on high number of iterations, but in practice this table does not store every PSO particle iterations value ; it just keep the best achieved PSO result on each successful novelty search iteration. The number of iterations of novelty search is much lesser than PSO search. Thats why this \textit{PSO\_archive} table is finite and easily traceable.

So, simultaneously all the leader particles may perform their PSO search( if not re-computing its position) and keep their best achieved position into the global \textit{PSO\_archive} table. We can adopt another approach where instead of storing the result after PSO search, we define a new parameter \textit{Gbest} where we keep the best achieved result obtained so far. Whenever, a Leader particle finishes its PSO search it will compare its newly-obtained result with \textit{Gbest}; if it is better than \textit{Gbest}, \textit{Gbest} will be updated with newly founded value and Leader particle will move on to the next novel location else no updation needed. We try both of them (\textit{PSO\_archive,Gbest}) in our experiment. \textit{PSO\_archive} gives us the trajectory of each Leader particles where as \textit{Gbest} gives us the exact optimum value and helps us to get rid of maintaining a global table and finding the result later.

Algorithm-(4) clearly explains the  Bare-Bones PSO algorithm.

\begin{figure}[h]
	\centering
	\begin{tikzpicture}
	
	\draw[blue,fill=LimeGreen!10](-1.5,1.5)node[black]{$P_1$} circle (1.5cm);
	
	\draw[VioletRed,fill=red](-1.2,1.1 )node[black]{} circle (0.50mm);
	\draw[VioletRed,fill=red](-0.5,1.3 )node[black]{} circle (0.50mm);
	\draw[VioletRed,fill=red](-1.3,0.8 )node[black]{} circle (0.50mm);	
	\draw[VioletRed,fill=red](-0.5,1.4 )node[black]{} circle (0.50mm);
	\draw[VioletRed,fill=red](-2.4,1.4 )node[black]{} circle (0.50mm);
	\draw[VioletRed,fill=red](-2.3,0.3 )node[black]{} circle (0.50mm);
	\draw[VioletRed,fill=red](-2.0,2.3 )node[black]{} circle (0.50mm);
	\draw[VioletRed,fill=red](-2.5,1.4 )node[black]{} circle (0.50mm);
	\draw[VioletRed,fill=red](-2.8,2.1 )node[black]{} circle (0.50mm);
	\draw[VioletRed,fill=red](-0.7,2.7 )node[black]{} circle (0.50mm);
	\draw[VioletRed,fill=red](-2.2,1.1 )node[black]{} circle (0.50mm);
	\draw[VioletRed,fill=red](-2.5,0.8 )node[black]{} circle (0.50mm);	
	\draw[VioletRed,fill=red](-0.7,2.5 )node[black]{} circle (0.50mm);
	
	\draw[->,thick] (-3.5,0)--(3.5,0) node[right]{$x$};
	\draw[->,thick] (0,-2.6)--(0,3.5) node[above]{$y$};
	
	\node [draw] at (-1.5,-3){Swarm particles inside the region of $P_1$};
	\draw[->,thick] (-1.4,-2.8)--(-1.2,1.1) node[]{};
	
	\end{tikzpicture}
	\caption{Inside of a region of leader particle $P_1$ where a group of particles are searching for global solution.}
\end{figure}
In the figure-(12), a typical region of leader particle $P_1$ is shown. Here $P_1$ calls Bare-Bone PSO and red dots are swarm particles generated randomly inside the region of $P_1$. They keep on searching that region until they converge to some solution.

\section{Algorithmic summary}
\label{sec:5}
The Novelty Search based Particle Swarm Optimization Algorithm(NSPSO) is represented in pseuducode as follows  -
see in page -(21-24).

There are 4 algorithms. \textit{NSPSO(), NC(),NS() and BBPSO()} All of them are quite self-explanatory.

\textit{NSPSO()} algorithm is the main algorithm. \textit{NSPSO()} algorithm calls Novelty Computation(NC()) algorithm for every iteration by every leader particles. After \textit{NC()} returns novelty value(N) to the \textit{NSPSO()}, \textit{NSPSO()} calls PSO search algorithm \textit{BBPSO()} if novelty value is grater than threshold value (\textit{$Th_N$}). If not then \textit{NSPSO()} calls \textit{RECOMP()} function which randomly select positions inside the application search region.

Algorithm \textit{NC()} calls Novelty Score Calculation algorithm \textit{NS()} every time it is required for every leader particles current position and their previous positions.

\begin{algorithm*}
	\caption{Novelty-Search-PSO Algorithm}
	\begin{algorithmic}[1]
		\Procedure{NSPSO()}{}
		\State $\textbf{\textit{-------------Initialization Phase----------------------------}}$
		\State $\text{Particle P} \gets \text{n}$
		\Comment {\textit{[n : Number of particles.]}}
		\State $\text{PositionTable PT(n)} \gets \text{NULL}$
		\Comment {\textit{[PT : Position Table holds the position after performing PSO search.]}}
		
		\State $\text{P(n)} \gets \text{Random Positions RAND()}$
		\Comment {\textit{[RAND() : Randomly define n number of positions for n particles.]}}
		\State $\text{F(n)} \gets \text{CALL FIT(P(n))}$
		\Comment {\textit{[FIT(P(n)) : Compute Fitness function for n particles based on applications.]}}
		\State $\text{PBEST(n,P)} \gets \text{NULL}$
		\Comment {\textit{[PBEST(n,P) : It holds best achieved fitness value and position after PSO search.]}}
		\State $\text{RT(n)} \gets \text{NULL}$
		\Comment {\textit{[RT(n) : Re computation Table.]}}
		
		\For {$ \text{i} \gets 1 $ to n$  $}
		\State $\text{N(i)} \gets \text{CALL NC(P(i),n)}$
		\Comment {\textit{[NC() : Novelty Computation algorithm.]}}
		\State $\text{PBEST(i,P(i))} \gets \text{(i,P(i))}$
		\Comment {\textit{[PBEST(i,P(i)) : It holds initial position of P(i).]}}
		
		\EndFor
		
		\State $\textbf{\textit{-------------------Main Algorithm Phase---------------------}}$

		\For {$ \text{itr} \gets 1 $ to \textit{MaxIterations}$ $}
		\Comment {\textit{[MaxIterations : Maximum number of iterations.]}}
		\For {$ \text{i} \gets 1 $ to n$ $}
		\If{N(i) $>=$ $Th{_n}$}
		\Comment {\textit{[$Th{_n}$ : Novelty Threshold value determined by the application.]}}
		\State $\text{bp} \gets \text{CALL PSO\_SEARCH(P(i))}$
		\Comment {\textit{[PSO\_SEARCH() : Calling PSO algorithm with particle P(i).]}}
		\If{FIT(bp) $\le$ FIT(PBEST(P(i))) }
		\State $\text{PBEST(i,P(i))} \gets \text{(i,bp)}$
		\Comment {\textit{[Update PBEST if new best position is better than old best position.]}}
		\EndIf
		
		\State $\text{PT(i)} \gets \text{P(i)}$
		\State $\text{P(i)} \gets \text{CALL RECOMP(P(i))}$
		\Comment {\textit{[RECOMP() : Redistribute particle to a new position.]}}
		\Else
		\State $\text{P(i)} \gets \text{CALL RECOMP(P(i))}$
		\EndIf 
		\EndFor
		
		\For {$ \text{i} \gets 1 $ to n$  $}
		\State $\text{N(i)} \gets \text{CALL NC(P(i),P(n))}$
		\Comment {\textit{[Calculate new novelty value for the new positions.]}}
		\EndFor
		
		\EndFor

		\EndProcedure
	\end{algorithmic}
\end{algorithm*}

\begin{algorithm*}
	\caption{Novelty Computation Algorithm}
	\begin{algorithmic}[1]
		\Procedure{NC(P(i),n)}{}
		
		\State $\text{N} \gets  \text{0}$
		\Comment {\textit{[N : Novelty value of particle P(i). ]}}
		\State $\text{j} \gets  \text{0}$	
		
		\State $\textbf{\textit{----Novelty Checking for Previously Searched Positions---}}$
		
		\While{PT(j) $\neq$ 0}
		\State $\text{tmp} \gets \text{CALL NS(P(i),PT(j))}$
		\Comment {\textit{[NS() : Novelty Score Calculating function.]}}
		
		\If{tmp $<$ $Th{_n}$}
		\State $\text{RT(i)} \gets \text{ "true"}$
		\Comment {\textit{[Send "recomputation" message to itself.]}}
		\State $\text{Break}$		
		\Comment {\textit{[This position is searched previously so leave immediately and recompute current position.]}}
		\EndIf 
		\State $\text{j} \gets \text{j+1}$
		
		\EndWhile
		
		\State $\textbf{\textit{----Novelty Checking for Current Positions----}}$	
		
		\State $\text{j} \gets  \text{0}$	
		\Comment {\textit{[Re initializing  local variable j.]}}
		
		\If{RT(i) $==$ "true"}
		\Comment {\textit{[Particle P(i)  receive "recomputation" message.]}}
		\State $\text{N} \gets 0$
		\Comment {\textit{[Forcefully making novelty score of P(i) very low.]}}
		
		\Else
		\While{j $\leq$ n}
		\State $\text{tmp} \gets \text{CALL NS(P(i),P(j))}$
		\Comment {\textit{[Calculate Novelty Score for all other particles.]}}
		\If{tmp $<$ $Th{_n}$}
		\State $\text{RT(j)} \gets \text{ "true"}$
		\Comment {\textit{[Send "recomputation" message to Other particles.]}}
		\EndIf 
		\State $\text{N} \gets \text{N+tmp}$		
		\Comment {\textit{[Summing Novelty Score for all the particles.]}}
		\State $\text{j} \gets \text{j+1}$
		\EndWhile
		\State $\text{N} \gets \text{N$/$(n-1)}$
		\Comment {\textit{[Averaging the Novelty Score for the particles P(i).]}}
		
		\EndIf 
		\State $\text{Return(N)}$
		
		\EndProcedure
	\end{algorithmic}
\end{algorithm*}

\begin{algorithm*}
	\caption{Novelty Score Calculation Algorithm}
	\begin{algorithmic}[1]
		\Procedure{NS(P(i),P(k))}{}
		\State $\text{ns} \gets  \text{0}$
		\State $\text{d} \gets  \text{0}$
		\State $\text{a} \gets  \text{P(i)}$
		\State $\text{b} \gets  \text{P(k)}$
		\Comment {\textit{[d: distance variable]}}
		\State $\text{d} \gets \ 	\sqrt{\sum_{i=1}^{dimension} \ (a_i^2 - b_i^2)}$		
		\Comment {\textit{[Computing Euclidean Distance for multidimensional points.]}}
		\If{d $\geq$ 2*r}
		\Comment {\textit{[r : radius value of the PSO region determined by applications.]}}
		\State $\text{ns} \gets  100$
		\Comment {\textit{[P(i) gets 100\% novelty.]}}
		\Else
		\State $\text{ns} \gets  (\frac{d}{2*r}) * 100$
		\Comment {\textit{[In all other cases P(i) gets "ns" novelty according to this formula.]}}
		
		\EndIf	
		
		\State $\text{Return (ns)}$

		\EndProcedure
	\end{algorithmic}
\end{algorithm*}
\begin{algorithm*}
	\caption{PSO\_SEARCH Algorithm}
	\begin{algorithmic}[1]
		\Procedure{BBPSO(P(i))}{}
		\State $\textbf{\textit{---We Use Bare Bones Particle Swarm Optimization Algorithm-----}}$
		\State $ \ P_n\ \gets \text{n}$
		\Comment {\textit{[Randomly initialize n particles inside the circular region centered by P(i).]}}
		\State $ \ f_n\ \gets \text{NULL}$
		\Comment {\textit{[$f_n$ : Holds fitness values of all the particles.]}}
		\For {$ \text{i} \gets 1 $ to n$ $}
		\State $ \ f_i\ \gets \text{CALL FIT}(P_i)$
		\Comment {\textit{Every particle P(i) computing its Fitness.]}}
		\State $ \ P_{Best_{i}}\ \gets P_i$
		\Comment {\textit{[In initialization phase Every Particle $P_i$'s position is $P_{Best}$.]}}
		\EndFor
		\State $ \ G_{Best}\ \gets { \{ P_i \mid f_i(P_i) \text{ is minimum} \} } $
		\Comment {\textit{[$G_{Best}$ holds best particles position among all $P_{Best}$.]}}
		
		\While{ $\ (G_{Best}\ $  $\leq$ $\ Th_f) \ $ or (itr $\leq$ \textit{MaxItr})}
		\Comment {\textit{[$Th_f$ : Fitness Threshold value determined by the application.]}}
		
		\For {$ \text{i} \gets 1 $ to n$ $}
		\State $m=\frac{G_{Best}+P_{Best}}{2}$
		\Comment {\textit{[m : Mean of $G_{Best}$ and $P_{Best}$.]}}
		\State $sd=\lvert G_{Best}-P_{Best} \rvert$
		\Comment {\textit{[sd : Standard Deviation of $G_{Best}$ and $P_{Best}$.]}}
		\State $\ newP_i\ \gets N(m,sd)$
		\Comment {\textit{[N(m,sd) : Gaussian Distribution of Mean and Standard Deviation.]}}
		\State $ \ newf_i\ \gets \text{CALL FIT}(newP_i)$
		\Comment {\textit{[$newf_i$ : New Fitness is Calculated based on $newP_i$.]}}
		\If{$newf_i$ $<$ $f_i$}
		\Comment {\textit{[Less then symbol($<$) used for minimization problems.]}}
		\State $ \ P_{Best_{i}}\ \gets newP_i$
		
		\EndIf 
		\State $ \ P_i\ \gets newP_i$
		\Comment {\textit{[New Position is updated.]}}
		
		\EndFor

		\State $ \ G_{Best}\ \gets { \{ P_i \mid f_i(P_i) \text{ is minimum} \} } $
		\Comment {\textit{[$G_{Best}$ updated based on new positions.]}}
		
		\EndWhile
		\If{$f(G_{Best})$ $<$ $Th_f$}
		\Comment {\textit{[If achieved $G_{Best}$ is better than fitness Threshold value return $G_{Best}$.]}}
		\State $\text{Return} (G_{Best})$
		
		\EndIf

		\EndProcedure
	\end{algorithmic}
\end{algorithm*}

\section{Experimental Result and Analysis}
\label{sec:6}
\subsection{Experimental Setting}
\label{sec:61}
The proposed Novelty Search based Particle Swarm Optimization algorithm approach is tested by MatLab 2015a on a 64 bit PC with Intel i5 processor with 3 GHz speed.

\subsection{Benchmark Setting}

For our testing purpose we choose mainly multimodal test functions and few unimodal problems. Our main objective is to check the performance of our algorithm \textbf{NSPSO} in complex environment where global optima is far from local optima and there are multiple local optima. We use two unimodal function and 14 multimodal benchmark functions \cite{Jq06},\cite{Xi14},\cite{Xi19},\cite{Co03},\cite{Ji19}.

All the functions are tested on 10 and 30 dimensions. 
Experimental test metrics are as follows --
\begin{enumerate}
	
	\item Running times : 25 times
	\item Population size : 40
	\item Maximal fitness evaluations (MaxFES) : 5000 * D * 60
	
\end{enumerate}
D is used as dimension of the problem. The above test metrics are used for all the comparative algorithms like CLPSO or ECLPSO but for NSPSO we use following metrics which are same as above except population size.

\begin{enumerate}
	
	\item Running times : 25 times
	\item Population size : 25
	\item Maximal fitness evaluations (MaxFES) : 5000 * D * 60
	\item Number of Leader Particles : 7
	
\end{enumerate}
We can divided the functions according to their properties into four groups : \textit{Unimodal Functions,Unrotated Multimodal Functions, Rotated Multimodal Function} and \textit{Composition Functions}.

The formulas and properties of all the test functions are described as follows --

\textit{Group A: Umimodal and Simple Multimodal Functions:}

1. Sphere Function
\begin{equation}
f_1(x) = \sum_{i=1}^{D} (x_i)^2
\end{equation}
2. Rosenbrock's Function
\begin{equation}
f_2(x) = \sum_{i=1}^{D-1} (100(x_i^2 -x_{i+1})^2 + (x_i -1)^2)
\end{equation}	

The first function is quite easy to solve and no need to discuss any detail here. But the second function is a multimodal function. We put this Rosenbrock's function here in Group A because it is a simple mutimodal function. Most of the algorithms which solve Sphere function can also solve Rosenbrock's function.
\textit{Group B:Unrotated Multimodal Functions:}

3. Ackley's Function
\begin{equation}
\begin{aligned}
f_3(x) = -20 exp \left(-0.2 \sqrt{\frac{1}{D}\sum_{i=1}^{D} x_i^2}\right) \\- exp \left(\frac{1}{D}\sum_{i=1}^{D} \cos(2\pi x_i) \right) +20 + e
\end{aligned}
\end{equation}	

4. Griewank's Function
\begin{equation}
\begin{aligned}
f_4(x) = \sum_{i=1}^{D} \frac{x_i^2}{4000} - \prod_{i=1}^{D}\cos \left( \frac{x_i}{\sqrt{i}} \right ) +1
\end{aligned}
\end{equation}	

5. Weierstrass's Function
\begin{equation}
\begin{aligned}
f_5(x) = \sum_{i=1}^{D} \left( \sum_{k=0}^{max} \left[a^k\cos(2 \pi b^k(x_i + 0.5)) \right] \right) \\ -D \sum_{k=0}^{max} \left[a^k\cos(2 \pi b^k. 0.5) \right],
\\a =0.5, b=3,max=20.
\end{aligned}
\end{equation}	

6. Rastrigin's Function
\begin{equation}
\begin{aligned}
f_6(x) = \sum_{i=1}^{D} \left(x_i^2 - 10\cos (2\pi x_i) +10\right)
\end{aligned}
\end{equation}	

7. Noncontinuous Rastrigin's Function
\begin{equation}
\begin{aligned}
f_7(x) = \sum_{i=1}^{D} \left(y_i^2 - 10\cos (2\pi y_i) +10\right)\\  
y_i = \begin{cases}
x_i, & |x_i| < \frac{1}{2}\\
\frac{round(2x_i)}{2}, & |x_i| \geq \frac{1}{2} \\
\text{for i} = 1,2,...D.
\end{cases} 
\end{aligned}
\end{equation}

8. Schwefel's Function
\begin{equation}
\begin{aligned}
f_8(x) = 418.9829 \times D - \sum_{i=1}^{D} x_i \sin\left( |x_i|^{\frac{1}{2}} \right )			
\end{aligned}
\end{equation}	
In this six multimodal functions, \textit{Ackley's function} has many local optima and one optimum basin. \textit{Griewank's Function} is more difficult problem in lower dimension than higher.  \textit{Weierstrass's Function} is continuous but also differentiable on some set of points.  \textit{Rastirgin function} is one of the best function because of high number of local optima. Our algorithm is perfect for this type of testing function. \textit{Noncontinuous Rastrigin's Function} contains same number of local optima and it is a good candidate for our test purpose. Another excellent function is \textit{Schwefel's Function} which has deep local optima and far form global optima.

\textit{Group C: Rotated Multimodal Functions :} For Rotation, we need an orthogonal matrix \textbf{M}. To create orthogonal matrix we use Salomon's method \cite{Sa19}. The original variable \textbf{x} is left multiplied by \textbf{M} to get the new rotated variable \textbf{y = M * x}. This \textbf{y} should be used to calculate fitness value f --
\begin{equation}
M = 
\begin{bmatrix}
m_{11} & m_{12} & ... & m_{1D}\\
m_{21} & m_22 & ... & m_{2D}\\
...  & ...  & ... & ... \\
m_{D1} & m_{D2} & ... & m_{DD}
\end{bmatrix}
\end{equation}
\textbf{x =} $\left[ x_1,x_2,...,x_D \right ]^T$ and \textbf{y =} $\left[ y_1,y_2,...,y_D \right ]^T$ then 
\begin{equation}
\begin{aligned}
y_i = m_{i1}x_1 + m_{i2}x_2 + ... + m_{iD}x_D,\\ i = 1,2,...D.
\end{aligned}
\end{equation}

9. Rotated Ackley's function
\begin{equation}
\begin{aligned}
f_9(x) = -20 exp \left(-0.2 \sqrt{\frac{1}{D}\sum_{i=1}^{D} y_i^2}\right) \\- exp \left(\frac{1}{D}\sum_{i=1}^{D} \cos(2\pi y_i) \right) +20 + e, \textbf{y = M * x}		
\end{aligned}
\end{equation}	

10. Rotated Griewank's function
\begin{equation}
\begin{aligned}
f_{10}(x) = \sum_{i=1}^{D} \frac{y_i^2}{4000} - \prod_{i=1}^{D}\cos \left( \frac{y_i}{\sqrt{i}} \right ) +1, \textbf{y = M * x}
\end{aligned}
\end{equation}	

11.Rotated Weierstrass's Function
\begin{equation}
\begin{aligned}
f_{11}(x) = \sum_{i=1}^{D} \left( \sum_{k=0}^{max} \left[a^k\cos(2 \pi b^k(y_i + 0.5)) \right] \right) \\ -D \sum_{k=0}^{max} \left[a^k\cos(2 \pi b^k . 0.5) \right],
\\a =0.5, b=3,max=20, \textbf{y = M * x}
\end{aligned}
\end{equation}	

12. Rotated Rastrigin's Function
\begin{equation}
\begin{aligned}
f_{12}(x) = \sum_{i=1}^{D} \left(y_i^2 - 10\cos (2\pi y_i) +10\right), \textbf{y = M * x}
\end{aligned}
\end{equation}	

13. Rotated Noncontinuous Rastrigin's Function
\begin{equation}
\begin{aligned}
f_{13}(x) = \sum_{i=1}^{D} \left(z_i^2 - 10\cos (2\pi z_i) +10\right)\\  
z_i = \begin{cases}
y_i, & |y_i| < \frac{1}{2}\\
\frac{round(2y_i)}{2}, & |y_i| \geq \frac{1}{2} \\
\text{for i} = 1,2,...D., \textbf{y = M * x}
\end{cases} 
\end{aligned}
\end{equation}

14. Rotated Schwefel's Function
\begin{equation}
\begin{aligned}
f_{14}(x) = 418.9829 \times D - \sum_{i=1}^{D} z_i\\
z_i = \begin{cases}
y_i \sin\left( |y_i|^{\frac{1}{2}} \right ), & |y_i| \text{if}\leq 500\\
0.001\left(|y_i| - 500\right)^2, & |y_i| > 500 \\
\text{for i} = 1,2,...D., \textbf{y = M * x}\\
\end{cases}		
\end{aligned}	
\end{equation}	
\textbf{y = y' + 420.96, y' = M * (x - 420.96)}	 

As we know Rotated Schwefel's Function has better results in the search range $[-500,500]^D$, so $z_i$ set in portion to the square distance between $y_i$ and the boundary.

\textit{Group D: \textbf{MMO} function:}
We choose three Multimodal Multiobjective Optimization functions from CEC19 \cite{Ji19}. We can choose other functions from latest CEC, like CEC2021 or CEC2022, but CEC19 gives us those functions which will be best tested for our algorithm. Multimodal multi-objective functions may have more than one \textit{Pareto Optimal Sets(\textbf{PSs})} corresponding to the same \textit{Pareto Front(\textbf{PF})}. Entire test suit contains test problems with different characters such as problems with different shape of PSs and PFs, coexistence of local and global PSs, scalable number of PSs, decision variables and objectives. For our test,\textbf{MMF1, MMF7 and MMF11} are chosen \cite{Ji19}.

1. \textit{\textbf{MMF1}} : This is Convex, Non-Liner function containing \textit{Pareto Optimal Front} and a scalable number of \textit{Pareto sets}. 	

\begin{equation}
\begin{aligned}
f_{15}(x) =	\begin{cases}
f_{15}^1 = |x_1 - 2|\\
f_{15}^2 = 1 - \sqrt{|x_1 - 2|}+ 2(x_2 - \sin(6\pi|x_1 - 2| \\+ \pi))^2	\\
\end{cases}	
\end{aligned}	
\end{equation}	

Its search space is - $x_1 \in [1,3],x_2 \in [-1,1]$. \\
Its global PSs are - 

\begin{align*}
\text{PS's are :}	\begin{cases}
x_1 = x_1\\
x_2 = \sin(6\pi|x_1 - 2| + \pi)
\end{cases}	
\end{align*}
where $1\leq x_1 \leq 3$,\\
Its global PFs are --
\begin{align*}
\text{PF's are :}	f_2 = 1 - \sqrt{f_1}
\end{align*}
where $0\leq f_1 \leq 1$.

2. \textit{\textbf{MMF7}} : This is Convex, Non-Liner function containing \textit{Pareto Optimal Front} but no pareto sets. 	

\begin{equation}
\begin{aligned}
f_{16}(x) =	\begin{cases}
f_{16}^1 = |x_1 - 2|\\
f_{16}^2 = 1 - \sqrt{|x_1 - 2|}+ \{x_2 - [0.3|x_2 - 2|^2 \cdot \\ \cos(24\pi|x_1 -2| + 4\pi) + 0.6|x_1 - 2|] \cdot \\ \sin(6\pi|x_1 - 2| + \pi)\}^2	\\
\end{cases}	
\end{aligned}	
\end{equation}	

Its search space is - $x_1 \in [1,3],x_2 \in [-1,1]$. \\
Its global PSs are - 

\begin{align*}
\text{PS's are :} \begin{cases}
x_1 = x_1\\
x_2 = [0.3|x_2 - 2|^2 \cdot \\ \cos(24\pi|x_1 -2| + 4\pi) + 0.6|x_1 - 2|] \cdot \\ \sin(6\pi|x_1 - 2| + \pi)
\end{cases}	
\end{align*}
where $1\leq x_1 \leq 3$.\\
Its global PFs are --
\begin{align*}
\text{PF's are :}
f_2 = 1 - \sqrt{f_1}
\end{align*}
where $0\leq f_1 \leq 1$.

3. \textit{\textbf{MMF11}} : This is Convex, Liner function containing \textit{Pareto Optimal Front}, scalable \textit{pareto sets} and also coexistence of global and local pareto sets. 	

\begin{equation}
\begin{aligned}
f_{17}(x) = Min	\begin{cases}
f_{17}^1 = x\\
f_{17}^2 = \frac{g(x_2)}{x_1}
\end{cases}	
\end{aligned}	
\end{equation}	
Where,
\begin{align*}
g(x) = 2 - exp \left[ -2log(2)\cdot \left(\frac{x - 0.1}{0.8} \right)^2\right]\cdot \sin^6(n_p\pi x)
\end{align*}
where $n_p$ is  total number of global  and local PSs.\\ 
Its search space is - $x_1 \in [0.1,1.1],x_2 \in [0.1,1.1]$. \\
Its global PSs are - 
\begin{equation}
\nonumber
x_2 = \frac{1}{2n_p}, x_1\in[0.1,1.1].
\end{equation}
its $i^{th}$ local PS is --
\begin{equation}
\nonumber
x_2 = \frac{1}{2n_p} + \frac{1}{n_p}\cdot(i -1), x_1\in[0.1,1.1].
\end{equation}
Where, i = 1,2,3,...,$n_p$.\\
its global PF is --

\begin{equation}
\nonumber
f_{17}^2 = \frac{g \left(\frac{1}{2n_p}\right)}{f_1}, f_1\in[0.1,1.1].
\end{equation}

Its local PF is --

\begin{equation}
\nonumber
f_{17}^2 = \frac{g \left(\frac{1}{2n_p} + \frac{1}{n_p}\cdot(i-1)\right)}{f_1}, f_1\in[0.1,1.1].
\end{equation}
where, i = 1,2,3,...,$n_p$.\\

\subsection{Parameter Settings for the involved PSO and other algorithms}
We will perform experiment on five state-of-the-art algorithms and compare their performance with our proposed algorithm. Among the five algorithms, three are particle swarm optimization algorithm which are combined with novelty search NS\cite{Bur20},NdPSO\cite{Gal15} and other one is PSONovE\cite{Ul21} and two algorithms are CLPSO and ECLPSO algorithm \cite{Xi14},\cite{Jq06} respectively which are different hybrid PSO methods best known for their high performance in multimodal test functions. The algorithms and their parameter settings are listed below  --
\begin{enumerate}
	\item Novelty Swarm (\textbf{NS}) \cite{Bur20}
	\item Novelty Driven PSO (\textbf{NdPSO}) \cite{Gal15}
	\item Novelty Search in PSO (\textbf{PSONovE}) \cite{Ul21}
	\item Comprehensive Learning PSO(\textbf{CLPSO}) \cite{Ji19}
	\item Extended Comprehensive Learning PSO(\textbf{ECLPSO}) \cite{Xi14}
	\item Novelty Search PSO (\textbf{NsPSO}) 
\end{enumerate}
We did not find too many algorithms where novelty search and PSO is combined. The NS, NDPSO and PSONovE methods are among them. So we choose them as the competitor algorithms. CLPSO and ECLPSO are the best so far algorithms on multi modal multi objective environment. In our experiment we perform sequential implementation. How the system will behave when parallel run will be implemented will be another interesting observation. Here in serial implementation, First leader particle perform its Novelty calculation, updates NS\_archive table. Then second leader particle and then third, so on.

Table 2 lists down global optima, search ranges and initialization range for the test functions.

\begin{table*}[h!]
	\caption{Global Optima, Search Range and Initialization Range of 17 Benchmark Functions.}
	\centering
	\renewcommand{\arraystretch}{2}
	\begin{tabular}{ ||c|c|c|c|c|| }
		\hline
		\textbf{Functions Number} &\textbf{x*} & \textbf{f(x*)} & \textbf{Search Range} & \textbf{Initialization Range}\\ 	
		\hline\hline
		\text{$f_1$} &${0}^D$ & 0  & $[-100,100]^D$ & $[-100,50]^D$\\  
		\hline
		\text{$f_2$} &${1}^D$ & 0  & $[-2.048,2.048]^D$ & $[-2.048,2.048]^D$\\  
		\hline
		\text{$f_3$} &${0}^D$ & 0  & $[-32.768,32.768]^D$ & $[-32.768,16]^D$\\  
		\hline
		\text{$f_4$} &${0}^D$ & 0  & $[-600,600]^D$ & $[600,200]^D$\\  
		\hline
		\text{$f_5$} &${0}^D$ & 0  & $[-0.5,0.5]^D$ & $[-0.5,0.2]^D$\\
		\hline  
		\text{$f_6$} &${0}^D$ & 0  & $[-5.12,5.12]^D$ & $[-5.12,2]^D$\\
		\hline
		\text{$f_7$} &${0}^D$ & 0  & $[-5.12,5.12]^D$ & $[-5.12,2]^D$\\
		\hline
		\text{$f_8$} &${420.96}^D$ & 0  & $[-500,500]^D$ & $[-500,500]^D$\\
		\hline
		\text{$f_9$} &${0}^D$ & 0  & $[-32.768,32.768]^D$ & $[-32.768,16]^D$\\
		\hline
		\text{$f_{10}$} &${0}^D$ & 0  & $[-600,600]^D$ & $[600,200]^D$\\ 
		\hline
		\text{$f_{11}$} &${0}^D$ & 0  & $[-0.5,0.5]^D$ & $[-0.5,0.2]^D$\\ 
		\hline
		\text{$f_{12}$} &${0}^D$ & 0  & $[-5.12,5.12]^D$ & $[-5.12,2]^D$\\ 
		\hline
		\text{$f_{13}$} &${0}^D$ & 0  & $[-5.12,5.12]^D$ & $[-5.12,2]^D$\\ 
		\hline
		\text{$f_{14}$} &${420.96}^D$ & 0  & $[-500,500]^D$ & $[-500,500]^D$\\ 
		\hline
		\text{$f_{15}$} &$[0,0]^D$ & [0,0]  & $[1,3]^D,[-1,1]^D$ & $[-100,100]^D$\\ 
		\hline
		\text{$f_{16}$} &$[0,0]^D$ & [0,0]  & $[1,3]^D,[-1,1]^D$ & $[-100,100]^D$\\ 
		\hline
		\text{$f_{17}$} &$[0,1]^D$ & [0,1]  & $[0.1,1.1]^D,[0.1,1.1]^D$ & $[-10,10]^D$\\ 
		\hline
	\end{tabular}
\end{table*}	
\subsection{Experimental Results and Discussions }

\subsubsection{Result for the 10-D Problems}

Table - 3 represents mean and variances of 25 runs of the five algorithms on seventeen test functions with D value  10. Best results are shown in bold. We also conducted \textit{\textbf{nonparametric Wilcoxon rank}} test to show that our \textbf{NsPSO}'s result is statistically different from other algorithms. There is a \textbf{h-value} in the table which indicates the result of \textit{t-}tests. The h =1 refers the performance of two algorithms are statistically different with 95\% certainty. The h = 0 means performance of two algorithms are not different.

\FloatBarrier
\begin{table*}[h]
	\caption{Comparative results for the 10 dimensional problems.}
	\centering
	\renewcommand{\arraystretch}{1.3}
	\resizebox{15cm}{!}{
		\begin{tabular}{||c|c|c|c|c||}
			
			\hline
			\textbf{Algorithms} &\textbf{Group A} & \textbf{Group A} & \textbf{Group B} & \textbf{Group B}\\ 	
			\hline\hline
			\textit{functions $-->$} & $f_1$ & $f_2$ & $f_3$ & $f_4$ \\  
			\hline
			\textit{CLPSO} & 5.15e-029 $\pm$ 2.16e-028 & \textbf{2.46e+000 $\pm$ 1.70e+000}  & 4.32e-014 $\pm$ 2.5e-01402 & 4.56e-003 $\pm$ 4.81e-003 \\  
			\hline
			\textit{ECLPSO} & \textbf{1.00e-97 $\pm$ 3.10e-96} & 2.70e+000 $\pm$ 1.50e+000  & 3.55e-015 $\pm$ 0 & \textbf{0 $\pm$ 0} \\
			\hline
			\textit{NdPSO} &9.32e-010 $\pm$ 5.12e-018 & 6.46e+000 $\pm$ 4.25e+000  & 7.32e-004 $\pm$ 4.5e-012 & 5.56e-001 $\pm$ 2.21e-001 \\  
			\hline
			\textit{NS} &8.17e-012 $\pm$ 2.16e-022 & 8.21e+000 $\pm$ 7.25e+000  & 7.23e-002 $\pm$ 6.5e-022 & 4.56e-001 $\pm$ 3.82e-001 \\ 
			\hline
			\textit{PSONovE} &1.87e-052 $\pm$ 1.26e-032 & 2.21e+000 $\pm$ 1.85e+000  & 6.53e-012 $\pm$ 3.1e-024 & 2.56e-021 $\pm$1.27e-001\\ 
			\hline
			\textit{NsPSO} &1.27e-089 $\pm$ 1.64e-053 & 3.33e+000 $\pm$ 1.78e+000  & \textbf{1.93e-022 $\pm$ 0} & \textbf{0 $\pm$ 0} \\ 
			\hline
			\textit{\textbf{h}} &1 & 1 & 1 & 0 \\ 
			\hline
			\hline
			\textbf{Algorithms} &\textbf{Group B} & \textbf{Group B} & \textbf{Group B} & \textbf{Group B}\\
			\hline 
			\textit{functions $-->$} & $f_5$ & $f_6$ & $f_7$ & $f_8$\\
			\hline
			\textit{CLPSO} &  \textbf{0 $\pm$ 0} & \textbf{ 0 $\pm$ 0}  &  \textbf{0 $\pm$ 0} &  \textbf{0 $\pm$ 0} \\
			\hline
			\textit{ECLPSO} & \textbf{0 $\pm$ 0} & \textbf{0 $\pm$ 0}  & \textbf{0 $\pm$ 0} & \textbf{0 $\pm$ 0} \\
			\hline
			\textit{NdPSO} &2.25e-021 $\pm$ 1.28e-031 & 4.36e-022 $\pm$ 2.25e-34  & 2.22e-024 $\pm$ 1.5e-22 & 2.36e-031 $\pm$ 3.31e-051 \\
			\hline
			\textit{NS} &1.95e-011 $\pm$ 0.98e-051 & 1.21e+20 $\pm$ 1.14e+000  & 2.23e-022 $\pm$ 4.5e-072 & 2.26e-021 $\pm$ 4.52e-061 \\ 
			\hline
			\textit{PSONovE} &0.25e-021 $\pm$ 0.38e-067 & 1.41e+000 $\pm$ 1.15e+000  & 1.53e-065 $\pm$ 1.12e-024 & 1.56e-041 $\pm$1.87e-021 \\ 
			\hline
			\textit{NsPSO} &\textbf{0 $\pm$ 0} & 0.13e-20 $\pm$ 1.78e+000  & \textbf{0 $\pm$ 0} & \textbf{0 $\pm$ 0} \\ 
			\hline
			\textit{\textbf{h}} &0 & 0 & 0 & 0\\ 
			\hline
			\hline
			\textbf{Algorithms} &\textbf{Group C} & \textbf{Group C} & \textbf{Group C} & \textbf{Group C}\\
			\hline 
			\textit{functions $-->$} & $f_9$ & $f_{10}$ & $f_{11}$ & $f_{12}$\\
			\hline
			\textit{CLPSO} &  3.65e-005 $\pm$ 1.57e-004 & 4.50e-002 $\pm$ 3.08e-002 & 3.72e-010 $\pm$ 4.40e-010 & 5.97e+000 $\pm$ 2.88e+000 \\
			\hline
			\textit{ECLPSO} & 3.55e-15 $\pm$ 0 & 2.22e-17 $\pm$ 4.53e-17  & 3.25e-025 $\pm$ 4.13e-20 & \textbf{2.27e+000 $\pm$ 4.47e+000} \\
			\hline
			\textit{NdPSO} &5.65e-017 $\pm$ 4.28e-011 & 2.46e-10 $\pm$ 4.25e-12  & 7.22e-024 $\pm$ 9.2e-022 & 7.57e-031 $\pm$ 9.22e-021 \\
			\hline
			\textit{NS} &6.95e-001 $\pm$ 529e-05 & 8.21e-25 $\pm$ 9.25e-053  & 7.75e-032 $\pm$ 7.89e-032 & 4.26e-031 $\pm$ 6.86e-041 \\ 
			\hline
			\textit{PSONovE} &2.65e-027 $\pm$ 2.38e-014 & 4.41e-55 $\pm$ 3.85e-30  & 2.23e-052 $\pm$ 3.5e-064 & 7.57e-071 $\pm$9.87e-08 \\ 
			\hline
			\textit{NsPSO} &\textbf{1.37e-025 $\pm$ 0} & \textbf{1.13e-022 $\pm$ 1.28e-20}  & \textbf{1.13e-023 $\pm$ 1.20e-13} & 3.15e+000 $\pm$ 5.3e+000 \\ 
			\hline
			\textit{\textbf{h}} &1 & 1 & 1 & 1\\ 
			\hline
			\hline
			\textbf{Algorithms} &\textbf{Group C} & \textbf{Group C} & \textbf{Group D} & \textbf{Group D}\\
			\hline 
			\textit{functions $-->$} & $f_{13}$ & $f_{14}$ & $f_{15}$ & $f_{16}$\\
			\hline
			\textit{CLPSO} &  5.44e+000 $\pm$ 1.39e+000 & 1.14e+002 $\pm$ 1.28e+002  & 1.22e-012 $\pm$ 2.3e-011 & 2.56e+000 $\pm$ 4.82e+000 \\
			\hline
			\textit{ECLPSO} & 2.01e+000 $\pm$ 3.78e+000 & \textbf{1.06e+003 $\pm$ 1.44e+002}  & 0.25e-025 $\pm$ 0 & 1.89e+000 $\pm$ 1.26e+000 \\
			\hline
			\textit{NdPSO} &4.25e+021 $\pm$ 5.28e+033 & 6.26e+002 $\pm$ 8.25e+002  & 7.22e-002 $\pm$ 4.3e-082 & 5.26e+001 $\pm$ 4.21e+001 \\
			\hline
			\textit{NS} &6.25e+011 $\pm$ 4.82e+21 & 3.21e+002 $\pm$ 5.26e+06  & 6.273e-002 $\pm$ 8.2e-075 & 4.99e+001 $\pm$ 3.32e+001 \\ 
			\hline
			\textit{PSONovE} &5.55e+051 $\pm$ 7.38e+027 & 2.88e+004 $\pm$ 1.35e+003  & 6.86e-011 $\pm$ 4.1e-034 & 2.88e+001 $\pm$5.77e+001 \\ 
			\hline
			\textit{NsPSO} &\textbf{1.16e+000 $\pm$ 4.72e+000} & 3.13e+000 $\pm$ 1.38e+001  & \textbf{0 $\pm$ 0} & \textbf{1.11e+000 $\pm$ 0.89e+000} \\ 
			\hline
			\textit{\textbf{h}} &1 & 1 & 1 & 1\\ 
			\hline
			\hline
			\cline{1-2}
			\textbf{Algorithms} &\textbf{Group D} & \multicolumn{3}{c||}{} \\ 
			\cline{1-2} 
			\textit{functions $-->$} &$f_{17}$ & \multicolumn{3}{c||}{} \\
			\cline{1-2}
			\textit{CLPSO} & 4.23e-023 $\pm$ 6.89e-45 &\multicolumn{3}{c||}{} \\
			\cline{1-2}
			\textit{ECLPSO} & 1.26e-24 $\pm$ 2.4e-42 &\multicolumn{3}{c||}{}  \\
			\cline{1-2}
			\textit{NdPSO} &11.25e-021 $\pm$ 13.22e-031 &\multicolumn{3}{c||}{} \\
			\cline{1-2}
			\textit{NS} &9.95e-011 $\pm$ 11.78e-051 &\multicolumn{3}{c||}{} \\ 
			\cline{1-2}
			\textit{PSONovE} &10.25e-051 $\pm$ 9.66e-027 &\multicolumn{3}{c||}{} \\ 
			\cline{1-2}
			\textit{NsPSO} &\textbf{1.19e-027 $\pm$ 0.72e-20} &\multicolumn{3}{c||}{}\\ 
			\cline{1-2}
			\textit{\textbf{h}} &1 &\multicolumn{3}{c||}{} \\ 
			\hline
			
		\end{tabular}
	}	
\end{table*}
\FloatBarrier
As we observe from the Table -(3), that for Group A unimodal functions our \textit{NsPSO} failed to provide best results. But certainly its comes second and third respectively in comparison with so far best multimodal PSO algorithms like CLPSO or ECLPSO. \textit{NsPSO} outperform existing other Novelty Search plus PSO hybrid algorithms. Our algorithm is designed for more robust and with more explorative power. Though the Rosenbrock's function is multimodal but due to novelty search our algorithm performs more exploration. Later we can improve by tuning parameters of novelty search such that it can perform much better on other unimodal functions.

For unrotated multimodal functions in Group B other than Rastrigin function our algorithm provides best results. Due to vast search region and more local optima novelty search provides better exploration than other algorithms. The \textit{h-value} is 0 whenever \textit{NsPSO} achieves same results as other algorithms. In case of function  $f_5$, though our algorithms beats other Novelty search PSO versions but it fails to converge into global optima.

In case of rotated multimodal functions in Group C, other than rotated rastrigin function, \textit{NsPSO} outperforms all algorithms. Rotation makes problems more difficult but instead of that difficulty \textit{NsPSO} shows quite promising performance in case of large local optima and global optima is far away. The more diverse the search region is, \textit{NsPSO} performs better than existing algorithms.

Again for more difficult valley in MMO problems in Group D, \textit{NsPSO} perform best. Here, we use very low number of PSO particles, where as \textit{NdPSO} or \textit{NS} or \textit{PSONovE} algorithms use higher values. Sometimes if we increase number of Leader particles \textit{NsPSO} gives even better results. It indicates that better tuning of the parameters of \textit{NsPSO} will give better performance. 

\subsubsection{Result for the 30-D Problems}
Same experiment is also done on 30 dimensional problems and their results are shown in Table - (4). In the case of Unimodal functions CLPSO beats everyone. \textit{NsPSO} shows good performance and highest among other novelty search based PSO methods. If we increase the number of iteration then there is a high chances that \textit{NsPSO} beats everyone. 

When dimension increases complexity of the problems simultaneously increases. But \textit{NsPSO} remains high performing and far better then other PSO methods.As we can see in some cases like function $f_6$ it fails to be the best yet it keeps it top performance.

In Group C rotated multimodal functions, \textit{NsPSO} shows remarkable results. In case of Rotated Rastrigin function NsPSO doesn't show  good performance similar to D=10 case. Also in case of Rotated Schwefel's function CLPSO shows better performance. Rotated Schwefel's function is complex function, and our algorithm is unable to beat CLPSO; but it shows competitive results in terms of other algorithms.

In one of the multimodal multiobjective function $f_{16}$ where there is no pareto sets and global optima is far away, our algorithm performs well in D=10, but fails to achieve best performance in D=30.

Comparing the performance of all the algorithms, we can see that ECLPSO and NsPSO performing far better than other hybrid PSO algorithms specially in multimodal complex situation where there are more valleys and global optima are distributed throughout search space.
\FloatBarrier
\begin{table*}[h!]
	\centering
	\caption{Comparative results for the 30 dimensional problems.}
	\renewcommand{\arraystretch}{1.5}
	\resizebox{15cm}{!}{
		\begin{tabular}{ ||c|c|c|c|c|| }
			\hline
			\textbf{Algorithms} &\textbf{Group A} & \textbf{Group A} & \textbf{Group B} & \textbf{Group B}\\ 	
			\hline\hline
			\textit{functions $-->$} & $f_1$ & $f_2$ & $f_3$ & $f_4$ \\  
			\hline
			\textit{CLPSO} & 4.46e-014 $\pm$ 1.73e-014 & \textbf{2.46e+000 $\pm$ 1.70e+000}  & \textbf{0 $\pm$ 0} & 3.146e-010 $\pm$ 4.64e-010 \\  
			\hline
			\textit{ECLPSO} & \textbf{2.02e-92 $\pm$ 1.22e-96} & 2.12e+000 $\pm$ 3.18e+000  & \textbf{0 $\pm$ 0} & 3.19e-010 $\pm$ 2.24e-010 \\
			\hline
			\textit{NdPSO} &9.27e-012 $\pm$ 3.12e-012 & 7.42e+000 $\pm$ 4.25e+000  & 7.22e-20 $\pm$ 5.5e-022 & 2.56e-001 $\pm$ 5.22e-001 \\  
			\hline
			\textit{NS} &2.13e-016 $\pm$ 4.16e-023 & 8.31e+002 $\pm$ 4.25e+002  & 6.23e-052 $\pm$ 4.5e-012 & 2.56e-03 $\pm$ 3.42e-002 \\ 
			\hline
			\textit{PSONovE} &4.27e-032 $\pm$ 6.86e-031 & 5.21e+002 $\pm$ 2.85e+001  & 6.78e-022 $\pm$ 3.1e-02 & 2.66e-011 $\pm$1.27e-001\\ 
			\hline
			\textit{NsPSO} & 2.87e-019 $\pm$ 2.54e-023 & 3.31e+000 $\pm$ 4.28e+000  & \textbf{0 $\pm$ 0} & \textbf{2.2e-015 $\pm$ 3.18e-012} \\ 
			\hline
			\textit{\textbf{h}} &1 & 1 & 0 & 1 \\ 
			\hline
			\hline
			\textbf{Algorithms} &\textbf{Group B} & \textbf{Group B} & \textbf{Group B} & \textbf{Group B}\\
			\hline 
			\textit{functions $-->$} & $f_5$ & $f_6$ & $f_7$ & $f_8$\\
			\hline
			\textit{CLPSO} &  3.45e-007 $\pm$ 1.94e-007 & 4.85e-010 $\pm$ 3.63e-010  & 4.36e-010 $\pm$ 2.44e-010 & 2.27e-012 $\pm$ 8.79e-13 \\
			\hline
			\textit{ECLPSO} & 4.33e-007 $\pm$ 1.89e-007 & \textbf{1.98e-011 $\pm$ 2.78e-10}  & 4.34e-10 $\pm$ 1.22e-010 & 1.11e-012 $\pm$ 3.89e-13 \\
			\hline
			\textit{NdPSO} &8.25e-011 $\pm$ 5.28e-011 & 6.36e-012 $\pm$ 4.25e-10  & 6.62e-014 $\pm$ 2.5e-12 & 3.33e-011 $\pm$ 4.31e-011 \\
			\hline
			\textit{NS} &6.95e-011 $\pm$ 5.58e-051 & 6.21e-10 $\pm$ 4.14e-000  & 4.23e-012 $\pm$ 4.5e-012 & 4.26e-011 $\pm$ 6.32e-011 \\ 
			\hline
			\textit{PSONovE} &7.15e-011 $\pm$ 6.28e-017 & 4.41e-012 $\pm$ 3.65e-012  & 7.53e-015 $\pm$ 4.12e-014 & 4.56e-011 $\pm$4.27e-011 \\ 
			\hline
			\textit{NsPSO} &\textbf{1.78e-009 $\pm$ 2.89e-009} & 2.13e-12 $\pm$ 3.75e-010  & \textbf{3.12e-12 $\pm$ 4.72e-12} & \textbf{0.45e-012 $\pm$ 0.11e-10} \\ 
			\hline
			\textit{\textbf{h}} &0 & 0 & 0 & 0\\ 
			\hline
			\hline
			\textbf{Algorithms} &\textbf{Group C} & \textbf{Group C} & \textbf{Group C} & \textbf{Group C}\\
			\hline 
			\textit{functions $-->$} & $f_9$ & $f_{10}$ & $f_{11}$ & $f_{12}$\\
			\hline
			\textit{CLPSO} &  3.43e-004 $\pm$ 1.91e-004 & 7.04e-010 $\pm$ 1.25e-011 & 3.07e+000 $\pm$ 1.61e+000 & 3.56e+001 $\pm$ 4.59e+000 \\
			\hline
			\textit{ECLPSO} & 3.15e-04 $\pm$ 0.78e-004 & 5.22e-11 $\pm$ 1.23e-11  & 2.25e+000 $\pm$ 1.13e+000 & \textbf{1.27e+000 $\pm$ 4.24e+000} \\
			\hline
			\textit{NdPSO} &5.25e-017 $\pm$ 7.29e-015 & 5.46e-10 $\pm$ 8.25e-12  & 7.67e+014 $\pm$ 9.32e+012 & 7.57e+011 $\pm$ 9.32e+011 \\
			\hline
			\textit{NS} &6.35e-001 $\pm$ 5.56e-11 & 8.21e-15 $\pm$ 6.25e-013  & 6.75e+012 $\pm$ 7.19e+012 & 5.26e+011 $\pm$ 6.66e+011 \\ 
			\hline
			\textit{PSONovE} &2.25e-017 $\pm$ 4.28e-016 & 4.61e-15 $\pm$ 5.15e-10  & 5.13e+012 $\pm$ 3.56e+014 & 7.17e+011 $\pm$1.87e+10 \\ 
			\hline
			\textit{NsPSO} &\textbf{2.14e-015 $\pm$ 1.14e-015} & \textbf{2.63e-012 $\pm$ 1.08e-11}  & \textbf{1.43e+000 $\pm$ 1.0e+000} & 2.15e+020 $\pm$ 3.2e+020 \\ 
			\hline
			\textit{\textbf{h}} &1 & 1 & 1 & 1\\ 
			\hline
			\hline
			\textbf{Algorithms} &\textbf{Group C} & \textbf{Group C} & \textbf{Group D} & \textbf{Group D}\\
			\hline 
			\textit{functions $-->$} & $f_{13}$ & $f_{14}$ & $f_{15}$ & $f_{16}$\\
			\hline
			\textit{CLPSO} &  3.77e+001 $\pm$ 5.56e+000 & \textbf{1.70e+003 $\pm$ 1.86e+002}  & 3.25e-011 $\pm$ 1.3e-010 & 5.56e+000 $\pm$ 4.2e+000 \\
			\hline
			\textit{ECLPSO} & 2.22e+001 $\pm$ 4.28e+001 & 2.61e+001 $\pm$ 2.42e+001  & 1.25e-015 $\pm$ 1.89e-015 & \textbf{1.19e+001 $\pm$ 3.46e+001} \\
			\hline
			\textit{NdPSO} &5.65e+011 $\pm$ 7.27e+013 & 6.92e+013 $\pm$ 9.75e+013  & 3.82e-012 $\pm$ 4.2e-012 & 5.39e+001 $\pm$ 6.27e+001 \\
			\hline
			\textit{NS} &6.89e+012 $\pm$ 5.22e+11 & 5.21e+012 $\pm$ 5.21e+012  & 6.214e-012 $\pm$ 8.7e-012 & 2.99e+002 $\pm$ 3.42e+002 \\ 
			\hline
			\textit{PSONovE} &7.25e+011 $\pm$ 7.28e+011 & 3.82e+001 $\pm$ 5.32e+001  & 2.82e-011 $\pm$ 5.11e-011 & 3.82e+002 $\pm$4.77e+002 \\ 
			\hline
			\textit{NsPSO} &\textbf{2.72e+001 $\pm$ 4.28e+001} & 2.53e+000 $\pm$ 3.38e+002  & \textbf{1.02e-010 $\pm$ 3.82e-010} & 2.71e+001 $\pm$ 3.29e+001 \\ 
			\hline
			\textit{\textbf{h}} &1 & 1 & 1 & 1\\ 
			\hline
			\hline
			\cline{1-2}
			\textbf{Algorithms} &\textbf{Group D} & \multicolumn{3}{c||}{} \\ 
			\cline{1-2} 
			\textit{functions $-->$} &$f_{17}$ & \multicolumn{3}{c||}{} \\
			\cline{1-2}
			\textit{CLPSO} &2.45e-013 $\pm$ 4.22e-13 &\multicolumn{3}{c||}{} \\
			\cline{1-2}
			\textit{ECLPSO} & 1.56e-14 $\pm$ 3.4e-14 &\multicolumn{3}{c||}{}  \\
			\cline{1-2}
			\textit{NdPSO} &10.25e-011 $\pm$ 11.22e-011 &\multicolumn{3}{c||}{} \\
			\cline{1-2}
			\textit{NS} &5.25e-011 $\pm$ 10.28e-011 &\multicolumn{3}{c||}{} \\ 
			\cline{1-2}
			\textit{PSONovE} &3.75e-011 $\pm$ 5.26e-011 &\multicolumn{3}{c||}{} \\ 
			\cline{1-2}
			\textit{NsPSO} &\textbf{0.89e-012 $\pm$ 1.72e-12} &\multicolumn{3}{c||}{}\\ 
			\cline{1-2}
			\textit{\textbf{h}} &1 &\multicolumn{3}{c||}{} \\ 
			\hline
		\end{tabular}
	}
\end{table*}
\FloatBarrier

\subsubsection{Experiment with more advanced benchmark set}

According to the latest benchmark study \cite{Ja22}, it has been observed that, commonly used evolutionary methods contain a center-biased operator which leads them to find optima in the center of the benchmark set quickly. Standard benchmark functions like Ackley, Griewank, Rosenbrock, Rastrigin or Schwefel has a design flaw, which is identified by CEC 2021 competition \cite{Mo20}. A large portion of this functions have optimum at zero vector or in the center of the feasible set. Although it is not a serious issue in terms of optimality of those functions but during comparison of various evolutionary methods which incorporate "Check-in-the-middle" procedure or a "Center-bias" operator produce huge advantage which is wrong.

To avoid this center-bias operator phenomena , we test our proposed method with new advanced benchmark set \cite{Ku22}.
The new advance benchmark functions contain a "zigzag" function which is constructed based on the following formula -

\begin{equation}
\begin{aligned}
z(x,k,m,\lambda) = \begin{cases}
1 - m + \frac{m}{\lambda}(|x|/k - \lfloor |x|/k \rfloor), if |x|/k - \lfloor |x|/k \rfloor \le \lambda, \\
1 - m + \frac{m}{1 - \lambda}(|x|/k - \lfloor |x|/k \rfloor), otherwise	\\
\end{cases}	
\end{aligned}
\end{equation}

We use four basic benchmark functions\cite{Ku22} described as follows.
 
\begin{equation}
\begin{aligned}
\phi_{1}(x,k,m,\lambda) = 3.10^{-9}|(x-40)(x-185)x(x+50)(x+180)|z(x,km,\lambda)+10|\sin(0.1x)
|\end{aligned}
\end{equation}

\begin{equation}
\begin{aligned}
\phi_{2}(x,k,m,\lambda) = \phi_{1}(\phi_{1}(x,k,m,\lambda),k,m,\lambda)
\end{aligned}
\end{equation}

\begin{equation}
\begin{aligned}
\phi_{3}(x,k,m,\lambda) = 3|\ln(1000|x|+1)|z(x,k,m,\lambda)+30-30|\cos(\frac{x}{10\pi})|
\end{aligned}
\end{equation}

\begin{equation}
\begin{aligned}
\phi_{4}(x,k,m,\lambda) = \phi_{3}(\phi_{3}(x,k,m,\lambda),k,m,\lambda)
\end{aligned}
\end{equation}
To obtain the benchmark functions for dimension D, we should use simple sum of the functions $\phi$ for the individual components but also modify the  inputs  by as shift vector s $\in [-100,100]^{D}$ and rotation/scaling matrix \textbf{M} : 
 		 \begin{equation}
 		 \begin{aligned}
 		 f_{j}(\textbf{x},k,m,\lambda) = \sum_{i=1}^{D}\phi_{j}(x_{j},k,m,\lambda) j =1,...4,\\
 		 F_{j}(\textbf{x},k,m,\lambda) = f_{j}(\textbf{M}_{j}(\textbf{x}-\textbf{s}_{j}),k,m,\lambda) j =1,...4
 		 \end{aligned}
 		 \end{equation}
we construct four basic benchmark functions (F1-F4) that combined the zigzag function with different multimodal functions, and that inherit the parameters from the zigzag function. Their construction starts with four 1-D functions $\phi_{1},..\phi_{4}$ which are formulated in equation -(28) to (32). For different $k,m, \lambda $ parameter values we follow \cite{Ja22}.
		 
We compare our method's performance with PSO(basic version), Differential Evolution(DE)algorithm, LSHADE and MadDE and HSES.Other than DE and PSO which are widely used canonical evolutionary algorithms, LSHADE is one of the most popular variants of adaptive DE and among the best performing algorithm in CEC Competitions in past few years. Multiple Adaptation DE Strategy(MadDE) is one of best performing algorithm in CEC'21 Competition. Hybrid Sampling Evolution Strategy (HSES), winner of the CEC'18 Competition. The results is shown in table - 5.

\FloatBarrier
\begin{table*}[h]
	\caption{Comparative results with advanced benchmark set with 30 different runs.}
	\centering
	\renewcommand{\arraystretch}{1.3}
	\resizebox{10cm}{!}{
		\begin{tabular}{||c |c  |c  |c|c|c|c|c||}
			
			\hline
			\textbf{ID} & & \textbf{DE} & \textbf{HSES} & \textbf{LSHADE} & \textbf{MadDE} & \textbf{PSO} & \textbf{NsPSO}\\ 	
			\hline\hline
			\textit{} &Min &4.7E-05 & 0.148 & 0 & 0 & 0.148 & 0 \\  
			\textit{} &Median &1.810 & 0.148 & 0.016 & 0.098 & 0.690 & 0.034\\
			\textit{F1} &Mean &1.700 & 0.182 & 0.045 & 0.139 & 0.783 & 0.317\\
			\textit{} &Max &2.978 & 0.533 & 0.227 & 0.377 & 1.822 & 0.091\\
			\textit{} &Std &0.877 & 0.075 & 0.056 & 0.100 & 0.505 & 0.102\\
			\hline
			\textit{} &Min &0 & 0.306 & 0.002 & 0.080 & 0.306 & 0.091 \\  
			\textit{} &Median &0.341 & 0.464 & 0.139 & 0.475 & 0.751 & 0.617\\
			\textit{F2} &Mean &1.250 & 0.507 & 0.132 & 0.485 & 0.993 & 0.827\\
			\textit{} &Max &19.712 & 0.918 & 0.268 & 0.997 & 2.598 & 1.231\\
			\textit{} &Std &3.616 & 0.119 & 0.087 & 0.163 & 0.606 & 0.127\\
			\hline	
			\textit{} &Min &0.158 & 0.079 & 0.254 & 0.158 & 1.066 & 0.145 \\  
			\textit{} &Median &0.582 & 0.237 & 0.482 & 0.380 & 2.587 & 0.017\\
			\textit{F3} &Mean &0.777 & 0.244 & 0.450& 0.353 & 2.770 & 0.123\\
			\textit{} &Max &2.816 & 0.464 & 0.649 & 0.533 & 6.899 & 0.664\\
			\textit{} &Std &0.620 & 0.092 & 0.088 & 0.094 & 1.282 & 0.317\\
			\hline
			\textit{} &Min &0.227 & 0.909 & 0.200 & 0.136 & 0.988 & 0.152 \\  
			\textit{} &Median &1.105 & 1.210 & 0.526 & 1.882 & 2.292 & 0.817\\
			\textit{F4} &Mean &1.110 & 1.221 & 0.527& 1.871 & 2.679 & 1.731\\
			\textit{} &Max &2.062 & 1.660 & 0.745 & 2.437 & 6.379 & 0.682\\
			\textit{} &Std &0.482 & 0.156 & 0.122 & 0.277 & 1.340 & 0.228\\
			\hline

		\end{tabular}
	}	
\end{table*}
\FloatBarrier

If we observe the table carefully we can find that in this difficult benchmark functions also our proposed methods works really well. Though it may not comes first in few occasions but if we compare it with basic PSO method, NsPSO's performance improves quite a lot. If we think through logically NsPSO's performance should not deviate drastically because we did not change a lot in PSO rather we provide BBPSO the benefit of high exploration capability of Novelty Search, which is observed in table -(5).

\section{Convergence - Divergence discussion of Novelty Search based PSO}

Most of the PSO variants discuss convergence analysis of their approach but in our method showing the convergence analysis is redundant as we did not change much on the analytical structure of PSO. Rather it is important to analyze how the divergence of novelty search benefits our algorithm and how both convergence of PSO and divergence of Novelty Search complement each other to provide much superior performance in various experimental functions.

The  main objective of novelty search is to quantify a new individual's divergence from past behaviors. To capture such divergence each individual is mapped into a point in \textbf{\textit{behavior space}}. That \textbf{\textit{behavior space}} is same as our defined \textbf{\textit{problem space}}. After an individual is mapped into the behavior space, its novelty is measured as the \textit{sparseness of its neighborhood} within that space.

We calculate this \textit{sparseness} by approximating a new individual’s average distance to its closest
neighbors among the current population. We directly map this \textit{Sparseness computation} in our problem space by employing \textit{Euclidean distance} of each individual \textit{Leader particles}. Here PSO particles are never driven by \textit{Objective functions}; rather the higher euclidean distance means higher novelty leads leader particles to explore more unknown search space. Sole purpose of the leader particles is to explore the problem space without worrying about convergence into a specific point. The gradient of search is simply towards what is \textit{new}, with no other explicit objective. Figure -(13) shows how this drive towards novelty differs from more traditional evolutionary computation like PSO. 

\begin{figure}[h]
	\centering
	\begin{tikzpicture}
	\begin{scope}
	\draw[blue,very thick](-8,0) rectangle (-5,3);
	\filldraw[red] (-7,1.5) circle (4pt); 
	\draw[blue, very thick] (-6,3) -- (-6,1);
	\draw[black, very thick] (-5.5,2.5) -- (-5.3,2.3);
	\draw[black, very thick] (-5.3,2.5) -- (-5.5,2.3);
	\draw[green, very thick,->](-6.8,1.7) -- (-6.2,2.2);
	
	\draw[blue,very thick](-2,0) rectangle (1,3);
	\filldraw[red] (-1,1.5) circle (4pt); 
	\draw[blue, very thick] (0,3) -- (0,1);
	\draw[black, very thick] (-1.5,2.0) circle (4pt); 
	\draw[black, very thick] (-1.1,2.3) circle (4pt); 
	\draw[black, very thick] (-0.7,2.2) circle (4pt); 
	\draw[black, very thick] (-0.3,1.9) circle (4pt); 
	\draw[green, very thick,->](-1.1,1.3) -- (-1.6,0.6);
	\draw[green, very thick,->](-0.8,1.3) -- (-0.5,0.6);
	\draw[green, very thick,->](-1.0,1.3) -- (-0.9,0.6);
	
	\end{scope}
	\end{tikzpicture}
	\caption{Fitness based search(on left), Novelty search (on right)}
\end{figure}

\subsection{Reward scheme in Novelty search in divergence scenario}

The gradient of improvement of a particular particle(behavior)(red dot) in both the novelty search and fitness based search is observed in the figure-(13). In fitness based situation, tendency to converge into the objective (black cross) restrict exploration of the search space. The solid vertical bar indicate behavioral constraints that the mutation makes it unable to cross. In novelty search the open circle represented previously explored behavior; in our case it is previously explored search area. The major observation here is that, novelty search's gradients diverge  while objective based search converge towards local optima. Novelty search guarantees that it will keep on moving as long as there is novel behavioral differences inside the search space. 

\subsection{Divergence discussion of novelty search}
In novelty search let us assume the existence of behavioral space by \textit{\textbf{{B}}}. For any individual \textit{x} at generation \textit{g} $\in$ [0,\textit{G}] in genotype space $\chi$ and a reference set \textit{\textbf{$R^{\textit{g}}$}} $\subset \chi$. Novelty search defines the \textit{novelty} in each generation g as -

\begin{equation}
\begin{aligned}
N^\textit{g} = \frac{1}{k}\sum_{i=0}^{k}d(\phi(x),\phi(u_i)
\end{aligned}
\end{equation}

where, $\phi: \chi \rightarrow \textbf{\textit{{B}}}$ is a mapping from parameter space to behavior space, and $u_i \in R^{\textit{g}}$ are such that $\phi(u_i)$ are the k-nearest neighbours of $\phi(x)$ in the set $\phi(R^{\textit{g}})$. This notation \textit{d(.)} in equation -(33) represents a metric function defined in \textbf{\textit{B}} which is assumed to be the \textbf{$l^{2}$} Euclidean norm. The reference set $R^{\textit{g}}$ is most often defined as $R^{\textit{g}} =A^g \cup P^g $.  The divergence of $R^{\textit{g}}$, is better explained in the following equation-

\begin{equation}
div(R^g) = \frac{\partial}{\partial x}(y) + \frac{\partial}{\partial y}(-x) = 0
\end{equation}

Equation -(34) shows the simplified divergence equation for 2D cases. in case of multidimensional problem space the partial derivatives increase accordingly. 
\begin{figure}[h]
\begin{center}
	\includegraphics[width=0.5\textwidth]{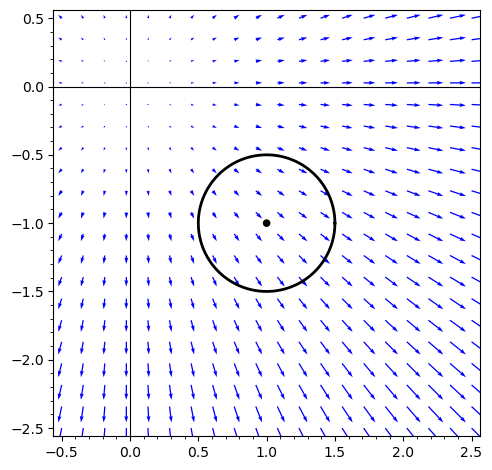}
	\caption{The divergence flow in the 2D problem space}
\end{center}
\end{figure}
Figure-(14) shows the conceptual description of divergence flow graph of novelty search. Black circled dot represent the individual's motion.

\subsection{Convergence discussion of BBPSO}
Velocity term eliminated in basic BBPSO, and two Gaussian random numbers are adopted. So update equation becomes 

\begin{equation}
x_{id}^{k+1} =  N(\mu,\sigma^2)
\end{equation}
where, $\mu  = (p_{gd}^k + p_{id}^k )/2$, $\sigma = |p_{gd}^k - p_{id}^k|$.

We can draw few conclusions here based on the study \cite{Fe08}, 

\textbf{Conclusion I:} $\forall k, \exists N$, when n $>$ N the canonical PSO can converge to BBSPO at any value.

\begin{equation}
\lim_{n>N}|x_{id}^{k+n} - \hat{x}_{id}^{k+n}| \le \epsilon, \forall k , \forall \epsilon,\exists N
\end{equation}

\textbf{Conclusion II:} Canonical PSO can be transfered to BBPSO with some simple modification.

\begin{equation}
\begin{cases}
\hat{\gamma}_{gd}^{k+n} \rightarrow \tilde{\gamma}_{gd}^{k+n} \sim N(\tilde{\mu},\tilde{\sigma^2}) \\
\hat{\gamma}_{id}^{k+n} \rightarrow \tilde{\gamma}_{id}^{k+n} \sim N(\tilde{\mu},\tilde{\sigma^2}) \\
\end{cases}       
\end{equation}
where, $\tilde{\mu} = \frac{1}{2}, \tilde{\sigma^2} = \frac{|p_{id}^{(k)} - p_{gd}^{(k)}|}{(p_{id}^{(k)})^2 + (p_{gd}^{(k)})^2 }$.

A simplification operation changes the the parameters $\hat{\gamma}_{gd}^{k+n}$ and $\hat{\gamma}_{id}^{k+n}$ by $\tilde{\gamma}_{gd}^{k+n}$ and $\tilde{\gamma}_{id}^{k+n}$, which are random values with Gaussian distribution $N(\tilde{\mu},\tilde{\sigma^2})$. Further statistical analysis gives us the following \textit{Expected} values-

\begin{align}
E[x_{id}^{(k+n)}] &= E[\tilde{\gamma}_{id}^{k+n} p_{id}^{(k)}] + E[\tilde{\gamma}_{gd}^{k+n} p_{gd}^{(k)}]\nonumber\\
&=p_{id}^{(k)}.E[\tilde{\gamma}_{id}^{k+n}] + p_{gd}^{(k)}.E[\tilde{\gamma}_{gd}^{k+n}]\nonumber\\
&=\tilde{\mu}.(p_{id}^{(k)} + p_{gd}^{(k)})
\end{align}

\begin{align}
D[x_{id}^{(k+n)}] &= D[\tilde{\gamma}_{id}^{k+n} p_{id}^{(k)} + \tilde{\gamma}_{gd}^{k+n} p_{gd}^{(k)}] \nonumber\\
&=(p_{id}^{(k)})^2.D[\tilde{\gamma}_{id}^{k+n}] + (p_{gd}^{(k)})^2.D[\tilde{\gamma}_{gd}^{k+n}]\nonumber\\
&=(\tilde{\sigma})^2.((p_{id}^{(k)})^2 + (p_{gd}^{(k)})^2)
\end{align}

According to the basic definition of basic bare bone pso, when $\tilde{\mu} = 0.5$ ,$\tilde{\sigma}^2 = \frac{|p_{id}^{(k)} - p_{gd}^{(k)}|}{(p_{id}^{(k)})^2 + (p_{gd}^{(k)})^2 } $. We can get the equation as -

\begin{equation}
x_{id}^{(k+n)} = \tilde{\gamma}_{gd}^{k+n} . p_{gd}^{(k)} + \tilde{\gamma}_{id}^{k+n}. p_{id}^{k}
\end{equation}

By setting $c_1$ and $c_2$ values 1.49, we achieve the mean and variance of (1 -  $\phi_{gd}^{(k+m)}$ -  $\phi_{id}^{(k+m)}$) as -
\begin{align}
E[1 -  \phi_{gd}^{(k+m)} -  \phi_{id}^{(k+m)}] = -0.495\\
D[1 -  \phi_{gd}^{(k+m)} -  \phi_{id}^{(k+m)}] = 0.371
\end{align}

A numerical experiment shows us the converging probability distribution as in figure-(15) and in figure-(16) -

\begin{figure}[h]
	\begin{center}
		\includegraphics[width=0.5\textwidth]{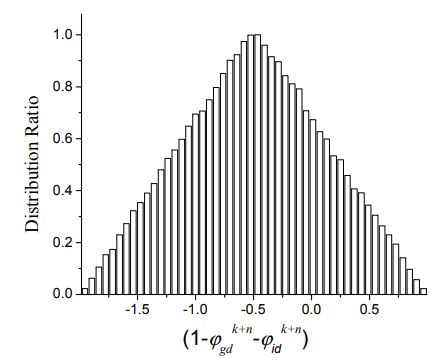}
		\caption{Numerical sampling of (1- $\phi_{gd}^{(k+m+1)} -  \phi_{id}^{(k+m+1)}$)}
	\end{center}
\end{figure}

\begin{figure}[h]
	\begin{center}
		\includegraphics[width=0.5\textwidth]{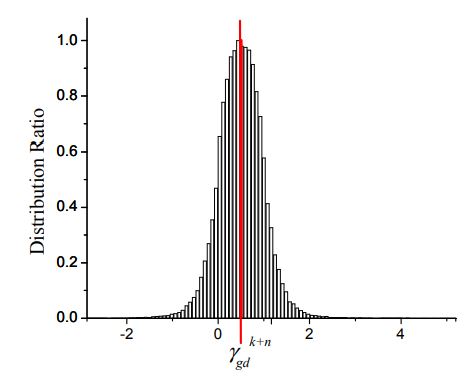}
		\caption{Numerical sampling of $\tilde{\gamma}_{gd}^{k+n}$}
	\end{center}
\end{figure}
\section{Conclusion}
\label{sec:7}
In this paper we propose a particle swarm optimization algorithm augmented with novelty search. We devise a method where PSO can successfully search entire area of the search region and never gets stuck in any local optima. The benefit of this algorithm is for more complex functions where multiple local and global optima exist and the solutions  are highly distributed throughout the search region. The present algorithm provides superior results.

Through our experiment we have seen that  \textbf{NsPSO} may not perform well in all kind of optimization problems. Though it is not designed to solve all tyeps of optimization problems but its main focus is to solve multi objective multi modal complex functions. 

The present algorithm is robust and dynamic because it has two parts one novelty search, which searches the novel area for searching and other is PSO algorithm. Now if any one wishes to use PSO variants other than BBPSO the present method will work successfully  and provide superior results.

\textbf{NsPSO} shows high performance on fourteen functions out of seventeen. It shows significant capability to search on complex domain and overcome the problem of locally stuck in false region.

\vskip 0.2in
\bibliography{bibtex}
\bibliographystyle{theapa}

\end{document}